\documentclass[conference]{IEEEtran}

\usepackage{times}

\usepackage[numbers]{natbib}
\usepackage{multicol}
\usepackage{xcolor}
\usepackage{amsmath, amssymb, yhmath}
\usepackage{graphics,graphicx,caption}
\usepackage[bookmarks=true]{hyperref}
\usepackage{booktabs}
\usepackage{array}
\usepackage{graphicx}
\usepackage{subcaption}
\usepackage[utf8]{inputenc}
\usepackage{amsmath, amsthm, amssymb}
\usepackage{bm}
\usepackage{graphicx}
\usepackage{booktabs}
\usepackage{dsfont}
\usepackage{xcolor}
\usepackage{url}
\usepackage{diagbox}

\newcommand{\policy}{Dense Transformation}

\begin{document}

\title{One Policy to Dress 
Them All: Learning to Dress People with Diverse Poses and Garments}



%
\author{\authorblockN{Yufei Wang,
Zhanyi Sun,
Zackory Erickson\authorrefmark{1}, 
David Held\authorrefmark{1}}
\authorblockA{
Robotics Institute, Carnegie Mellon University\\
\{yufeiw2, zhanyis, zerickso, dheld\}@andrew.cmu.edu \\
*Equal Advising
}
}

\makeatletter
\let\@oldmaketitle\@maketitle
\renewcommand{\@maketitle}{\@oldmaketitle
\centering
  \includegraphics[width=\linewidth]{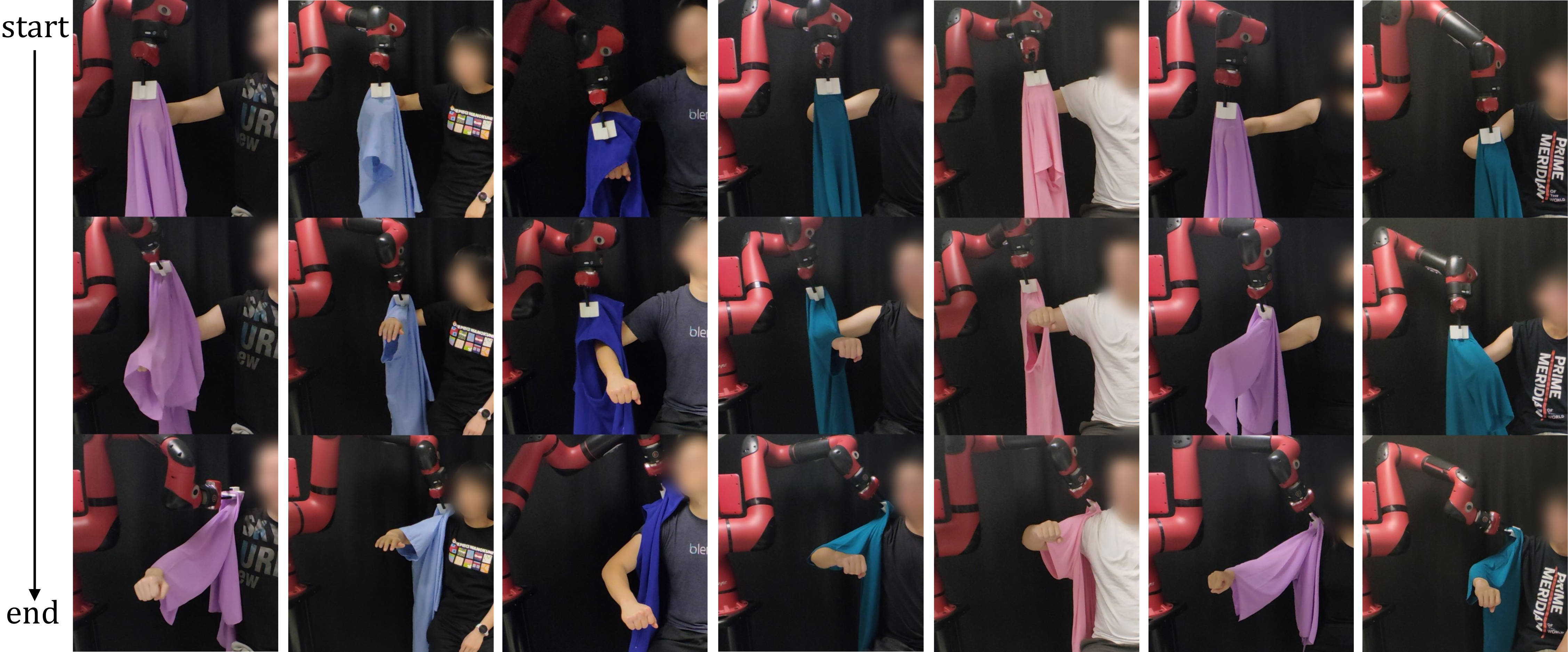}
  \captionof{figure}{
  We develop a robot-assisted dressing system based on a single learned policy that is able to dress different people with diverse poses and garments. Each column shows snapshots from a trajectory of our policy. We learn the dressing policy using reinforcement learning with point cloud observations to generalize to diverse garments. We use policy distillation to combine multiple policies that each work for a small range of arm poses into a single policy that works for a large variety of arm poses. 
  }
  \label{fig:intro_example}
  \vspace{-0.15in}
  }
\makeatother

\maketitle
\addtocounter{figure}{-1}

\maketitle

\begin{abstract}
Robot-assisted dressing could benefit the lives of many people such as older adults and individuals with disabilities.
Despite such potential, robot-assisted dressing remains a challenging task for robotics as it involves complex manipulation of deformable cloth in 3D space. 
Many prior works aim to solve the robot-assisted dressing task, but they make certain assumptions such as a fixed garment and a fixed arm pose that limit their ability to generalize.
In this work, we develop a robot-assisted dressing system that is able to dress different garments on people with diverse poses from partial point cloud observations, based on a learned policy. 
We show that with proper design of the policy architecture and Q function, reinforcement learning (RL) can be used to learn effective policies with partial point cloud observations that work well for dressing diverse garments. 
We further leverage policy distillation to combine multiple policies trained on different ranges of human arm poses into a single policy that works over a wide range of different arm poses.
We conduct comprehensive real-world evaluations of our system with 510 dressing trials in a human study with 17 participants with different arm poses and dressed garments. Our system is able to dress 86\% of the length of the participants' arms on average. Videos can be found on our project webpage~\footnote{\url{https://sites.google.com/view/one-policy-dress}}.

\end{abstract}

\section{Introduction}

Robot-assisted dressing could benefit the lives of many people. Dressing is a core activity of daily living. 
A 2016 study by the National Center for Health Statistics~\cite{harris2019long} shows that 92\% of all residents in nursing facilities and at-home care patients require assistance with dressing. 
Such needs will likely keep growing due to aging populations. 
A robot-assisted dressing system could not only improve life quality for older adults and individuals with disabilities by helping them maintain independence and privacy, but also help mitigate the growing shortage of nursing staff, and provide some much needed respite for caregivers.

Despite its potential societal impact, robot-assisted dressing remains a challenging task for robotics for the following reasons. 
As in many deformable object manipulation tasks, there is no compact state representation of cloth, and the dynamics of garments are non-linear and complex~\cite{zhu2022challenges}. Compared to table-top cloth manipulation tasks such as folding and smoothing that can be solved via pick-and-place actions, the dressing task demands more dexterous manipulation actions in 3D space.
Furthermore, people with disabilities and older adults usually have a limited range of motion and thus the dressing robot needs to generalize to a diverse range of arm poses. Finally, there are many different types of garments with varying geometries and material properties, e.g., short-sleeve hospital gowns and long-sleeve jackets. It is non-trivial for a robot to be able to assist in dressing all these garments, since dressing a slender long-sleeve elastic jacket might require very different end-effector trajectories compared with dressing a wide, short-sleeve non-elastic hospital gown.
All these factors render robot-assisted dressing a very challenging problem. 

There have been many prior works that investigate robot-assisted dressing, yet they make certain assumptions that limit their ability to generalize.
Most prior works~\cite{erickson2018deep, kapusta2019personalized, gao2016iterative, erickson2019multidimensional, ildefonso2021exploiting, zhang2019probabilistic, gao2015user} assume that the robot is dressing a known fixed garment
and thus does not generalize to dressing diverse garments. 
Other prior works simplify the assistive dressing problem by assuming that the person holds a single pose~\cite{erickson2018deep, qie2022cross}, or can move his arm into a pose that has high probability of dressing success~\cite{kapusta2019personalized, ildefonso2021exploiting, gao2016iterative}. We aim to improve upon these prior works to build a dressing system that can generalize to diverse garments and human poses. 

In this work, we learn a single policy that is able to dress diverse garments on different people holding diverse poses from partial point cloud observations using a single depth camera. 
We use point clouds as input to our policy in order to represent and generalize to different garments and human arm poses. 
We show that with proper design of the policy and Q-function architectures,
reinforcement learning (RL) can be used to learn effective policies with partial point cloud observations that work well for dressing diverse garments. 
We further leverage policy distillation to combine multiple policies trained on different ranges of human arm poses into a single one that works over a wide variety of different poses.
Finally, for robust sim2real transfer, we employ ``guided domain randomization'', which trains a policy with randomized observations by imitating policies trained without any randomization. The final domain-randomized policy can be reliably transferred to a real robot manipulator and dress real people. We conducted a human study with 17 participants, and performed 510 total dressing trials with different arm poses and garments. 
On average, our system is able to dress 86\% of the length of the participants' arms, and achieves a statistically significant difference in responses as compared to alternative baselines in a 7-point Likert item.
We also provide a comprehensive analysis of our dressing system's performance in the human study to understand its strength and weakness.

In summary, we make the following contributions: 
\begin{itemize}
    \item We develop a robot-assisted dressing system based on a learned policy with partial point cloud observations that generalizes to different people, arm poses, and garments.
    \item We show the effectiveness of policy distillation to increase the effective working range of the policy.
    \item  We perform comprehensive real-world evaluations of our system on a manikin, as well as with 510 dressing trials in a human study with 17 participants of varying arm poses and dressed garments. 
\end{itemize}

\section{Related Work}
\subsection{Robot Assisted Dressing}
There have been many prior works in robot-assisted dressing~\cite{zhang2022learning} which can be categorized as follows. 
Some works make the assumption that the person being dressed is collaborative in performing the dressing task~\cite{kapusta2019personalized, clegg2020learning}, while others do not~\cite{erickson2018deep, gao2016iterative}. We also do not assume a collaborative human in this work. Instead, we simplify the problem by assuming that the human is holding a fixed pose throughout the dressing procedure.

A large body of works focus on user modeling and building a personalized dressing plan for each human participant~\cite{kapusta2019personalized, zhang2017personalized, gao2016iterative, gao2015user, canal2019adapting, jevtic2018personalized, zhang2019probabilistic, li2021provably}. In contrast to these prior works, we do not focus on user modeling and aim to learn a single policy that is able to generalize to diverse poses and body sizes of different human participants. We leave the integration of user modeling for future work.

Another line of work studies haptic perception and simulation during robot-assisted dressing~\cite{erickson2017does, yu2017haptic, kapusta2016data, wang2022visual}. In contrast, we demonstrate how a robot can perform dressing assistance using only partial point clouds from a single off-the-shelf depth camera. 
Some prior works focus on learning where along a garment to grasp in preparation for dressing~\cite{qie2022cross, zhang2020learning}. In this work we make the assumption that the garment has already been grasped by the robot, but our method can be combined with these prior work to remove this assumption.

Reinforcement learning has been used in some prior works for dressing backpacks~\cite{ildefonso2021exploiting} or hospital gowns~\cite{clegg2020learning}. However, they assume a fixed garment and thus the policy has limited generalization towards different garments. Our policy can generalize to dressing diverse garments as we use point cloud as the garment representation.

\subsection{Robotic Deformable Object Manipulation}
Deformable object manipulation has long been a core task for robotics. It is challenging due to the complicated dynamics of deformables, high-dimensional state representation, and perception complexities such as self-occlusion. Many prior deformable object manipulation tasks such as cloth smoothing~\cite{lin2022learning, ha2022flingbot, xu2022dextairity, wu2019learning}, cloth folding~\cite{weng2022fabricflownet, avigal2022speedfolding, lee2021learning}, bedding manipulation~\cite{seita2019deep, puthuveetil2022bodies}, dough rolling~\cite{lin2022planning}, rope reconfiguration~\cite{seita2021learning, lin2021softgym}, and bag manipulation~\cite{bahety2022bag, seita2021learning, chen2022autobag} focus on the 2D table-top setting. Many of these prior cloth manipulation tasks can be solved by simple pick-and-place actions or the use of other pre-defined motion primitives~\cite{lin2022learning, weng2022fabricflownet, seita2019deep, seita2021learning, wu2019learning, hoque2022visuospatial, ha2022flingbot, xu2022dextairity}. Our work looks at the challenging problem of dressing assistance, which involves manipulating a deformable garment suspended in 3D space with complex closed-loop actions during physical human-robot interaction.

\subsection{Policy Learning for  Manipulation From Point Clouds}
There has been much recent work that aims to learn robotic manipulation policies from point cloud observations. 
Some of these works propose new algorithms for imitation learning from point clouds for grasping and manipulating tools or articulated objects~\cite{wu2022learning, seita2022toolflownet, eisner2022flowbot3d, pan2022tax}. 
Some recent works explore applying RL with point cloud observations for grasping or dexterous hand manipulation with rigid objects~\cite{wang2022goal, qin2022dexpoint, liu2022frame, huang2021generalization}.  
Our work differs from these as we apply RL from point cloud observations for deformable cloth manipulation.

\begin{figure*}[t]
    \centering
    \includegraphics[width=\textwidth]{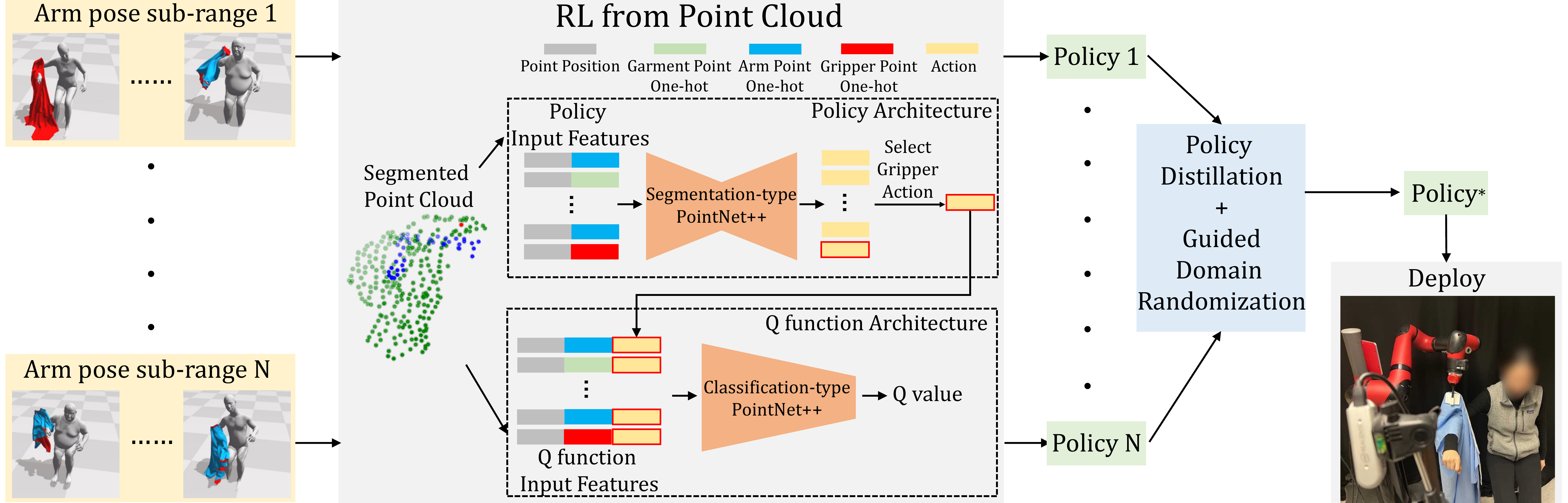}
    \caption{An overview of our proposed robot-assisted dressing system. (Left) In simulation, we divide the diverse arm pose range into multiple sub-ranges to ease policy learning. (Middle) We propose a new policy architecture and a corresponding Q function architecture for reinforcement learning from partial point cloud observations to learn effective dressing policies on each of the divided pose sub-ranges. (Right) We then leverage policy distillation to combine policies working on different pose sub-ranges into a single policy that works for a diverse range of poses. We also perform guided domain randomization for sim2real transfer, and we deploy the distilled policy to a real-world human study dressing real people.
    }
    \label{fig:system}
    \vspace{-0.6cm}
\end{figure*}

\section{Task Definition and Assumptions}
As shown in Fig.~\ref{fig:intro_example}, the task we study in this paper is single-arm dressing, where the goal is to fully dress the sleeve of a garment onto a person's arm, and the task is considered to be complete when the sleeve of the garment covers the person's shoulder.
The person can hold different arm poses before the dressing starts. The arm pose is defined by three joint angles $\phi = [\phi_1, \phi_2, \phi_3] \in \mathbb{R}^3$, where $\phi_1$ is the lifting and lowering angle of the shoulder, $\phi_2$ is the inwards-outwards bending angle of the elbow towards the body, and $\phi_3$ is the lifting and lowering angle of the elbow (Fig.~\ref{fig:pose_and_garment} illustrates these joint angles).
A depth camera is used to record depth images of the scene, and partial point clouds $P$ can be computed from the depth image. The ultimate goal of the paper is to learn a policy $\pi$ to dress diverse garments $g \in G$ for people holding a diverse range of arm poses $\phi \in \Phi$. At each time step, the policy takes as input the point cloud $P$ and outputs an action $a = \pi(P)$ as the delta transformation for the robot end-effector.
The garment set $G$ we consider includes hospital gowns and common everyday garments such as vests and cardigans with different sleeve lengths, geometries, and materials. 
The set of arm poses $\Phi$ is specified by the min and max values for each joint angle:  $\Phi = \{[\phi_1, \phi_2, \phi_3] ~|~ \phi_i \in [\phi^{min}_i, \phi^{max}_i], i = 1, 2, 3\}$. Fig.~\ref{fig:pose_and_garment} shows the garments and arm poses in $\Phi$ that we test in the real-world human study. 

We make two assumptions for the dressing task. 
First, we assume that the robot has already grasped a part of the garment around the opening of the garment shoulder in preparation for dressing, since grasping is not the focus of our work. Besides,  there has been some prior work that learn where along a garment to grasp for dressing~\cite{qie2022cross, zhang2020learning}, and our system can be combined with these prior work to remove this assumption.
This assumption has been used in prior work as well~\cite{erickson2018deep, kapusta2019personalized}. 
Second, we assume that the person holds the pose static during the dressing process. This assumption helps address the visual occlusion of the
arm caused by the cloth during the dressing process;  with this
assumption we can obtain the static arm point cloud before the dressing starts. 
This assumption has also been validated in other studies with participants with impairments~\cite{kapusta2019personalized}.
We leave for future work adapting to human motion
during the dressing process.

\vspace{-0.05in}
\section{Method}
\label{sec:Method}
\vspace{-0.1in}

\subsection{System Overview} 
As illustrated in Fig.~\ref{fig:system}, our robot-assisted dressing system consists of the following components. 
We first use reinforcement learning (RL) to learn the dressing policy $\pi$ in a physics-based simulation by formulating the robot assisted dressing problem as a Partially Observable Markov Decision Process (POMDP). We design a special 
policy network architecture for efficient training of effective RL policies from partial point cloud observations.
As it is hard to train a single policy that generalizes to a diverse range of arm poses $\Phi$, we divide the arm pose range into multiple sub-ranges $\{\Phi^{sub}\}_{i=1}^N$ and train a policy $\pi_i$ on each of them. We then use policy distillation to combine these different policies into a single policy $\pi^*$ that works for the wide range of arm poses $\Phi$.
For robust sim2real transfer, we further train the policy $\pi^*$ with domain randomization in the policy observation, with a  behaviour cloning loss to imitate policies that are trained without any randomization, a procedure which we name as ``guided domain randomization''.
Finally, we deploy the domain-randomized policy to a real robot that successfully dresses different participants with diverse poses and garments in a human study.

\begin{figure}
    \centering
    \includegraphics[width=\columnwidth]{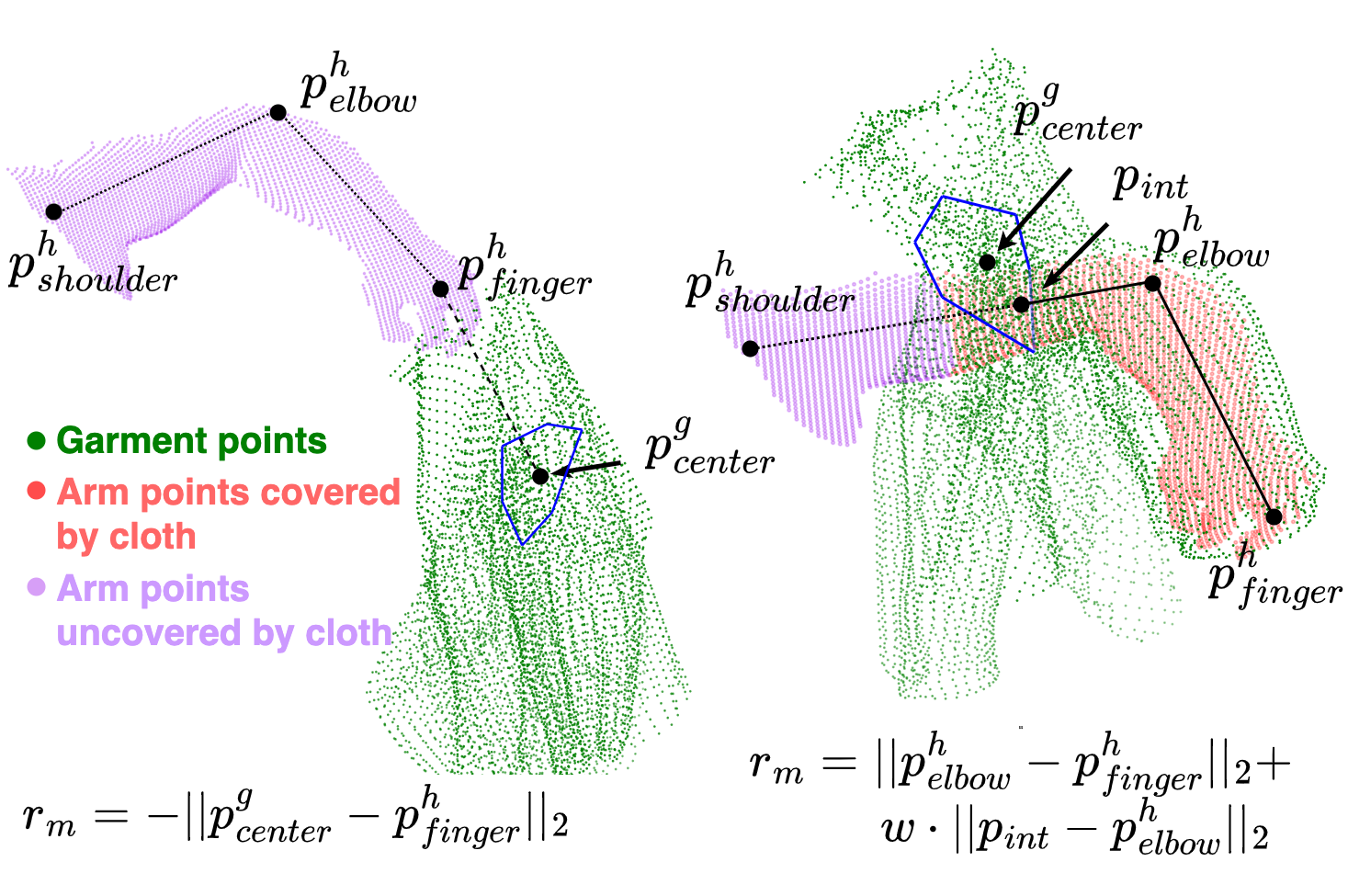}
    \vspace{-0.3in}
    \caption{Illustration of the main reward for the dressing task. (Left) The garment is not dressed onto the person's arm. (Right) The garment is dressed onto the person's arm. 
    }
    \label{fig:reward}
    \vspace{-0.25in}
  
\end{figure}
\vspace{-0.05in}
\subsection{Learning to Dress with Reinforcement Learning}
\label{sec:RL}
We use reinforcement learning in simulation to train policies for the dressing task. We formulate the robot-assisted dressing task as a POMDP $\langle S, A, O, R, T, U \rangle$, where $S$ is the state space, $A$ is the action space, $O$ is the observation space, $R$ is the reward function, $T$ is the transition dynamics, and $U$ is the measurement function that generates the observation from the state. We now detail the design of the core components of the POMDP as follows.

\textbf{Observation Space $O$}: 
Due to the lack of a compact state representation for cloth, the garment naturally requires a high-dimensional representation such as an image, mesh, or point cloud.
To facilitate sim2real transfer, 
we use the partial point cloud $P$ of the dressing scene computed from a depth camera as the policy observation. 
We perform cropping on the point cloud 
to only keep points that belong to the right arm of the human, denoted as $P^h$, and those that belong to the garment, denoted as $P^g$. In simulation we can easily perform such cropping using the ground-truth simulator information; see Section~\ref{sec:human_study_procedure} for details on how we perform such cropping in the real world. 
We further add a single point at the position of the robot end-effector to the point cloud to represent the robot gripper, denoted as $P^r$. The policy observation is then the concatenation of the arm, garment, and robot gripper points: $o = [P^h; P^g; P^r]$. See the middle part of Fig.~\ref{fig:system} for an example of the segmented point cloud. 

\textbf{Action Space $A$:} The action $a \in SE(3)$ is the delta transformation of the robot end-effector. We represent the delta transformation as a 6D vector, where the first 3 components are the delta translation, and the second 3 components describe the delta rotation using axis-angle. We set the roll rotation to be 0 in the action as it is not necessary for the dressing task. 

\textbf{Reward Function $R$}: 
Fig.~\ref{fig:reward} illustrates the main reward $r_m$ we use for the dressing task, which measures the progress of the task. The dressing task can be divided to two phases. 
Before the garment is dressed onto the person's forearm, the reward $r_m$ is the negative distance from the garment shoulder opening center $p^g_{center}$ to the finger of the person $p^h_{finger}$: $r_m = - ||p^g_{center} - p^h_{finger}||_2$.
After the garment is dressed onto the person's forearm, we compute the reward $r_m$ based on the distance the garment has been dressed onto the person's arm. To compute the dressed distance, we first approximate the opening of the garment shoulder as a hexagon (shown in blue in Fig.~\ref{fig:reward}). Then, we represent the person's forearm as a line that connects the elbow point $p^h_{elbow}$ and the finger point $p^h_{finger}$, and we represent the upper arm as another line that connects the shoulder point $p^h_{shoulder}$ and the elbow point $p^h_{elbow}$. Next, the intersection point $p_{int}$ between the garment shoulder opening hexagon and the two arm lines are computed. If the intersection point $p_{int}$ is on the forearm, $r_m$ equals the dressed distance, which is the distance between the intersection point and the person's finger point: $r_m = ||p_{int} -  p^h_{finger}||_2$. If the intersection point $p_{int}$ is on the upper arm, the dressed distance is the length of the forearm plus the distance between the intersection point and the elbow point, and $r_m$ is a weighted combination of these two: $r_m = || p^h_{elbow} -  p^h_{finger}||_2 + w \cdot || p_{int} -  p^h_{elbow}||_2$. We set $w=5$ to encourage the policy to turn at the elbow and dress  the upper arm.

In addition to the main reward term that measures the task progression, we also have three additional reward terms. The first is a force penalty $r_f$ that prevents the robot from applying too much force through the garment to the person. The second is a contact penalty $r_c$ that prevents the robot end-effector from moving too close to the person. 
The last reward term is a deviation penalty $r_d$ that discourages the garment center from moving too far away from the arm. 
The full reward is given by:
$r = r_m + r_f + r_c + r_d$.
More details of how these terms are computed can be found in Appendix Section A. 
Note that the reward function is only available in simulation, as it requires access to the ground-truth garment and human mesh information, which is non-trivial to estimate in the real world.

\textbf{Policy and Q function Architecture:}
Most prior works~\cite{liu2022frame, qin2022dexpoint, wang2022goal} that train RL policies with point cloud observation use a classification-type PointNet-like~\cite{qi2017pointnet, qi2017pointnet++} network architecture for the policy, which encodes the whole partial point cloud to a single action vector. There have been some recent works showing that instead of compressing the whole point cloud into a single action vector, inferring the action from a dense output leads to better performance~\cite{zeng2018learning, wu2020spatial, zeng2021transporter, wu2021spatial, seita2022toolflownet, eisner2022flowbot3d, ha2022flingbot, canberk2022cloth}. We follow the dense action representation idea and propose a new policy architecture named \textit{\policy} for reinforcement learning from point clouds.

As shown in Fig.~\ref{fig:system},
the input to the policy is a segmented point cloud $P = \{p_i\}_{i=1}^M$ of size $M$, which contains the garment points $P^g$, human right arm points $P^h$, and a single point at the robot end-effector position representing the robot gripper $P^r$. The features for each point $p_i$ include its 3D position, 
and a 3-dimensional one-hot vector
indicating the class of the point, i.e., whether the point belongs to the garment, the human arm, or the robot gripper. 
Instead of using a classification-type neural network architecture (e.g., a classification-type PointNet++) that compresses the whole point cloud into a single action vector, the \policy~policy uses a segmentation-type neural network architecture (e.g., a segmentation-type PointNet++) that outputs a dense per-point action vector $\{a_i\}_{i=1}^M$ (see Fig.~\ref{fig:system}). Among these $M$ action vectors, we only execute the action vector $a^*$ corresponding to the gripper point $P^r$, i.e, we select the action $a^* = a_j$, where $j$ is the index of the gripper point. We could alternatively use a classification-type PointNet++ that encodes the whole point cloud to a single action, but we find that using a segmentation-type network leads to slightly better performance (though the difference is relatively small).

For the Q function, given the current point cloud $P$ and the action sampled from the \policy~policy $a^* \sim \pi(P)$, we concatenate $a^*$ as an additional feature to every point $p_i$ in the input point cloud, so the feature of each point includes its 3D position, the one-hot vector of the class type, and the action $a^*$.  We then use a classification-type neural network architecture (i.e., classification-type PointNet++) to output a scalar Q value. This Q function architecture has also been used in prior work~\cite{wang2022goal}, although with a different policy architecture.  
We compare to other ways of representing the Q function in experiments and show that this one works the best. We use SAC~\cite{haarnoja2018soft} as the underlying RL algorithm, and we use PointNet++~\cite{qi2017pointnet++} for the policy and Q function network.

\vspace{-0.03in}
\subsection{Generalization to Diverse Human Poses  via Policy Distillation}
\vspace{-0.05in}
\label{sec:policy_distillation}
Learning a dressing policy that works on a diverse range of arm poses can be viewed as a multi-task learning problem, where each small range of poses can be considered as an individual task. 
The same holds true for learning a policy that works for diverse garments -- each garment can be treated as an individual task. 
In our experiments, we find it is possible to learn a single universal policy for a diverse set of garments, and the performance is similar to that of learning an individual policy for each garment. 
However, we find it challenging to learn a single policy that works well for a diverse range of arm poses, possibly due to imbalanced learning speed for different tasks (e.g., some poses are easier to learn compared with others), conflicting gradients from different tasks (the desired trajectory for one arm pose might contradict to another), and other potential issues. 
Inspired by~\cite{rusu2015policy,schmitt2018kickstarting}, we use policy distillation to address this issue.

\textbf{Policy Distillation.}
Given a set of policies that were each trained  for a single task, policy distillation~\cite{rusu2015policy} can be used to combine the set of policies into a single policy that works for all of the tasks. It has been shown that this often outperforms directly training a single policy for all of the tasks.
 In our case, we train individual policies each for dressing a human arm pose sub-range; we then distill these policies into a single policy that works for the full diverse range of arm poses $\Phi$.

To employ policy distillation, we first need to decompose the diverse arm pose range $\Phi$ into smaller ranges $\{\Phi^{sub}_i\}_{i=1}^N$ where RL can be directly used to learn effective policies on these smaller pose ranges. As aforementioned, the arm pose range $\Phi$ is specified by min and max values for each of the three joint angles  $\Phi = \{[\phi_1, \phi_2, \phi_3] ~|~ \phi_i \in [\phi^{min}_i, \phi^{max}_i], i = 1, 2, 3\}$. We perform the decomposition by dividing each joint angle range 
$[\phi^{min}_i, \phi^{max}_i]$ into a number of smaller intervals. For example, we can uniformly divide the range $[\phi^{min}_i, \phi^{max}_i]$ to $\Phi_i^j = [\phi^{min}_i + (j-1)\delta, \phi^{min}_i + j \delta], j = 1, ..., L$ ,
where $\delta = (\phi^{max}_i - \phi^{min}_i) / L$. This will then result in $L^3$ sub-ranges $\{\Phi^{sub} = \Phi_1^j \times \Phi_2^k \times \Phi_3^l \}_{j=1, k=1, l=1}^L$.

After we decompose the diverse arm pose range $\Phi$ into $N$ smaller ranges, we train $N$ ``teacher'' policies $\{\pi^t_i\}_{i=1}^N$, one for each pose sub-range, using RL as described in Section~\ref{sec:RL}. We then distill these $N$ teacher policies $\{\pi^t_i\}_{i=1}^N$ into a single ``student'' policy $\pi_s$ by training the student policy with a combination of the RL loss and a policy distillation loss, shown in Eq.~\ref{eq:loss}.

Specifically, let $\theta_s$ denote the parameters of the student policy $\pi_s$, let $\theta_t^i$ denote the parameters of the $i^{th}$ teacher policy $\pi_i^t$, and let $\mathcal{N}(\mu_\theta(o), \sigma_\theta(o)) = \pi_\theta(o)$  denote the Gaussian distribution of the action output by policy $\pi_\theta$ on observation $o$. Given a batch of samples collected by the student policy $\pi_{s}$, $\{o_n, a_n, r_n, o'_n\}_{n=1}^B$, we use the following loss to train $\pi_{s}$:
\begin{equation}
\begin{split}
        L(\theta_s) =& L_{SAC}(\theta_s) +  \beta \sum_{i=1}^N L_{distill}(\theta_s, \theta_t^i),
        \label{eq:loss}
\end{split}
\end{equation}
where $L_{SAC}(\theta_s)$ is the standard SAC training loss, $L_{distill}$ is the policy distillation loss (described below), and $\beta$ weighs the two terms. The policy distillation loss computes the distance between the student action distribution and the teacher action distribution, and thus the student learns to imitate the teacher's behaviors by minimizing such a loss.
Specifically, we compute the Earth Mover's distance between the action distribution of the student and the teacher policy:
\begin{equation}
\scriptsize
    \begin{split}
        L_{distill}(\theta_s, \theta_t^i) =&
        \sum_{n=1}^B \Big(\mu_{\theta_s}(o_n)-\mu_{\theta_t^i}(o_n)\Big)^2 
        + \Big(\sqrt{\sigma_{\theta_s}(o_n)}-\sqrt{\sigma_{\theta_t^i}(o_n)}\Big)^2
    \end{split}
    \label{eq:earth_mover}
\end{equation}
The loss in Eq.~\ref{eq:loss} has also been used in prior work~\cite{schmitt2018kickstarting}; however, they used the KL divergence instead of Earth Mover's distance for the distillation loss. Our experiments indicate that earth-mover distance leads to significantly improved performance. We set $\beta= 0.01$ in our experiments.

\vspace{-0.05in}
\subsection{Guided Domain Randomization Learning for Sim2real Transfer}
\vspace{-0.05in}
\label{sec:guided_distillation}
We find that there is a huge difference between the simulated garment point cloud and the real-world garment point cloud, due to large variations in simulation vs. real garment geometries, and also since the simulator does not perfectly model the dynamics of garment deformations in the real world.
To make the observation more aligned between simulation and the real world, and to make the policy robust to the observation difference, we add randomizations to the policy observation (described at the end of this section). Let $o$ denote the original non-randomized point cloud observation, and let $\tilde{o}$ denote the randomized observation. Naively training the policy with randomized observation $\tilde{o}$ will usually fail or lead to degraded performance, since the randomizations make policy learning more difficult. 
To mitigate this issue, we propose ``guided domain randomization'', where we first train teacher policies $\pi_i^t$ without any domain randomization; we then distill the teachers into a student
policy $\pi_s$ trained with domain randomization on the observation. 
Let $\theta_s$ denote the parameters of the observation randomized policy $\pi_s$. To train the student policy $\pi_s$, we run the student to collect a batch of data that stores both the randomized and non-randomized observation $\{\tilde{o}_n, o_n, a_n, r_n, \tilde{o}'_n, o'_n\}_{n=1}^B$, and train it as follow:
\begin{equation*}
\scriptsize
    \begin{split}
        &L(\theta_s) =\tilde{L}_{SAC}(\theta_s) + \beta \sum_{i=1}^N\tilde{L}_{distill}(\theta_s, \theta_t^i), \\
        &  \tilde{L}_{distill}(\theta_s, \theta_t^i) =
        \sum_{n=1}^B \Big(\mu_{\theta_s}(\tilde{o}_n)-\mu_{\theta_t^i}(o_n)\Big)^2 
        + \Big(\sqrt{\sigma_{\theta_s}(\tilde{o}_n)}-\sqrt{\sigma_{\theta_t^i}(o_n)}\Big)^2,
    \end{split}
\end{equation*}
where $\tilde{L}_{SAC}(\theta_s)$ is the SAC loss with running the policy $\pi_s$ on the randomized observations, and $\tilde{L}_{distill}(\theta_s, \theta_t^i)$ is the loss of imitating the teacher policy $\pi^t_i$ trained without domain randomization. Importantly, note that the student receives the randomized observation $\tilde{o}_n$ whereas the teacher receives the non-randomized observation $o_n$.
For the observation randomization, we perform random cropping, dropping, erosion, and dilation on the cloth point cloud, and we add random noise to the robot gripper position.  
More details on the observation randomization can be found in Appendix Section A.

\section{Simulation Experiments}

\subsection{Experimental Setup}
We train our policies using Softgym~\cite{lin2021softgym} based on the NVIDIA Flex simulator. 
We use SMPL-X~\cite{SMPL-X:2019} to generate human meshes of different body sizes and arm poses. To generate random poses, the shoulder joint $\phi_1$ is uniformly sampled from $[-20, 30]$, the inwards-outwards elbow joint $\phi_2$ is uniformly sampled from $[-20, 20]$, and the upwards-downwards elbow joint $\phi_3$ is uniformly sampled from $[-20, 30]$. We decompose the arm pose range $\Phi$ into $N=27$ regions for policy distillation, by dividing each joint angle range into $3$ intervals. 
We randomly pick 5 garments for training from the Cloth3D dataset~\cite{bertiche2020cloth3d}: a hospital gown and 4 cardigans. The selected garments have different geometries such as varying sleeve lengths and widths. 
We randomly generate 50 human poses for each of the 27 pose sub-ranges, resulting in a total of $27 \cdot 50 \cdot 5 = 6750$ different configurations. Among the 50 poses for each sub-range, we use 45 poses for training and 5 poses for evaluation. 
At each training episode of SAC, we randomly sample one garment out of the five for training.  
The simulated dressing environment with the person holding different poses and with different garments is shown in Fig.~\ref{fig:system}. More details of the simulation experimental setup can be found in Appendix Section B. 

\subsection{Baselines and Ablations}
We compare the following RL algorithms for learning the dressing policy from partial point cloud observations:
    \textbf{\policy~policy (ours)} is our proposed method, which is described in Section~\ref{sec:Method}. 
 \textbf{\policy~policy + latent Q function:} This baseline uses the same policy as our method. 
 For the Q function, instead of concatenating the action to the input point cloud as an additional feature, this baseline first uses a PointNet++ to encode the point cloud observation into a latent vector, and then concatenates the action with the latent vector. An MLP then takes as input the concatenated latent vector and outputs a scalar Q value.    
\textbf{Direct Vector} uses a classification PointNet++ policy to encode the input point cloud to a single action vector. It uses a similar Q function network architecture as our method, where the action is concatenated to every point in the point cloud as an additional feature.
\textbf{TD-MPC~\cite{hansen2022temporal}}, a state-of-the-art model based RL algorithm. 
\textbf{Deep Haptic MPC} \cite{erickson2018deep}, which learns a force prediction model based on end-effector measurements, and combines it with MPC to plan forward actions that minimizes the predicted force during dressing. The actions in MPC are sampled based on a ``moving forward'' heuristic. Note that this method does not use any visual information of the arm or the garment.

We also compare different ways of enabling the policy to generalize to diverse poses. 
    \textbf{Policy Distillation} is our proposed method, as described in Section~\ref{sec:policy_distillation}.
    \textbf{Policy Distillation (KL)} replaces the
    Earth Mover's distance in Equation~\eqref{eq:earth_mover} with KL divergence.
     \textbf{PCGrad~\cite{yu2020gradient}} is a multi-task learning algorithm that balances gradients from different tasks;  in our case, a ``task" is defined as training on a different pose sub-range.
      \textbf{No Distillation} trains a single policy on the entire pose range, without any distillation.
      \textbf{Heuristic Motion Planning} finds a collision-free robot end-effector path along the human arm based on some heuristically designed constraints. 
More implementation details of these baselines can be found in Appendix Section B. 

In simulation, we use the \textit{upper arm dressed ratio} as the evaluation metric. The ratio is computed between the dressed upper arm distance and the actual upper arm length: $\frac{||p_{int} - p_{elbow}^h||_2}{||p^h_{shoulder} - p_{elbow}^h||_2}$; see Section~\ref{sec:RL} and Fig.~\ref{fig:reward} for how these points are defined. This ratio is upper bounded by 1.

\begin{table}[t]
\scriptsize
\centering
\begin{tabular}{c|c|c|c|c|c}
\toprule
               & \multicolumn{1}{c|}{\begin{tabular}[c]{@{}c@{}}Dense \\ Transform \\ (Ours)\end{tabular}} & \multicolumn{1}{c|}{\begin{tabular}[c]{@{}c@{}}Dense \\ Transform\\ Latent Q\end{tabular}} & \multicolumn{1}{c|}{\begin{tabular}[c]{@{}c@{}}Direct \\ Vector\end{tabular}} & \multicolumn{1}{c|}{TD-MPC} & 
               \multicolumn{1}{c}{\begin{tabular}[c]{@{}c@{}}Deep Haptic \\ MPC \end{tabular}}
               \\ \hline
 1 & $\mathbf{0.68 \pm 0.05}$ & $0.55 \pm 0.05$ &  $0.63 \pm 0.07$ &  $0.04 \pm 0.04$ &  $0.31 \pm 0.08$ \\ \hline
2   & $\mathbf{0.55 \pm 0.06}$  & $0.42 \pm 0.08$  &  $0.52 \pm 0.09$ &    $0.00 \pm 0.00 $   &  $0.11 \pm 0.02$  \\ \hline
3   &  $\mathbf{0.75 \pm 0.04}$ & $0.68 \pm 0.07$ &  $0.73 \pm 0.02$ &   $0.03 \pm 0.05$  &  $0.37 \pm 0.02$ \\ 
\bottomrule
\end{tabular}
\caption{Upper arm dressed ratio of different policies on 3 pose sub-ranges. Results are averaged across 3 seeds.
}
\label{tab:RL_performance}
    \vspace{-0.1in}
\end{table}

\begin{table}[t]
\centering
\scriptsize
\begin{tabular}{c|c|c|c|c}
\toprule
\multicolumn{1}{c|}{\begin{tabular}[c]{@{}c@{}}Policy  \\ Distillation \\ (Ours) \end{tabular}} & \multicolumn{1}{c|}{\begin{tabular}[c]{@{}c@{}}Policy \\ Distillation \\ (KL) \end{tabular}} & 
\multicolumn{1}{c|}{PCGrad} & 
\multicolumn{1}{c|}{\begin{tabular}[c]{@{}c@{}}No \\ Distillation\end{tabular}} & 
\multicolumn{1}{c}{\begin{tabular}[c]{@{}c@{}}Heuristic \\ Motion \\ Planning\end{tabular}}
\\ 
\hline       
$\mathbf{0.68 \pm 0.012}$ & $0.45 \pm 0.010$ & $0.37 \pm 0.063$ & $0.34 \pm 0.10$ & 0.32
\\
\bottomrule
\end{tabular}
\caption{Upper arm dressed ratio of different ways to enable the policy to generalize to diverse poses. 
Results are averaged across 3 seeds. The Heuristic Motion Planning baseline does not have a standard deviation because there is no learning or randomness in this method. 
}
\label{tab:distillation_performance}
    \vspace{-0.2in}
\end{table}

\subsection{Does the \policy~policy perform better than the alternative baselines for learning the dressing policy?}
We first compare the performance of different RL algorithms.
We randomly select 3 different sub pose-ranges, and for each method, we train a separate policy on these 3 pose sub-ranges to compare their performances.
We train each method for 1e6 steps, evaluate at each 10K steps, and report the maximum performance. 
The results are shown in Table~\ref{tab:RL_performance} and averaged across 3 seeds. 
As shown, our proposed method \policy~policy achieves the best performances on all 3 pose sub-regions, although Direct Vector has similar performance. Using a ``Latent Q" architecture leads to much worse performance.  We find that TD-MPC~\cite{hansen2022temporal} does not work at all for this dressing task, possibly since the algorithm is originally designed and tested on image and state observations instead of point clouds, and also since it might be hard to learn a good latent dynamics model for deformable garments.  
Deep Haptic MPC also does not work well. The original paper tested only one arm pose and garment, with no visual information of the garment or human pose, potentially explaining its low performance in our setting with diverse arm poses and garments.
Based on this result, we use \policy~as our base RL algorithm for the following experiments. 

\subsection{Does policy distillation improve the ability of our method to generalize to diverse poses?}
We now compare the performance of different methods for learning a single policy on all 27 pose sub-ranges. 
We train all methods long enough until they converge (as different methods require different numbers of environment steps to converge), and report the maximum achieved performance.
Results are averaged across 3 seeds.
As shown in Table~\ref{tab:distillation_performance}, policy distillation with the Earth Mover's distance outperforms all other baselines. The baseline ``No Distillation'' and ``PCGrad~\cite{yu2020gradient}'' both perform poorly, showing the difficulty of directly learning a single policy across diverse pose ranges without distillation.
The performance of the ``Heuristic Motion Planning'' baseline is also low. We find it performs well when there is no sharp bending at the shoulder and the elbow, and worse when the bends are sharp, aligning with findings in prior work~\cite{kapusta2019personalized}.
Interestingly, we find that using Earth Mover's distance noticeably outperforms KL divergence, even though KL divergence seems to be the most common choice for policy distillation in the literature~\cite{rusu2015policy, schmitt2018kickstarting, parisotto2015actor}.

\begin{figure}[t]
    \centering
    \includegraphics[width=.5\columnwidth]{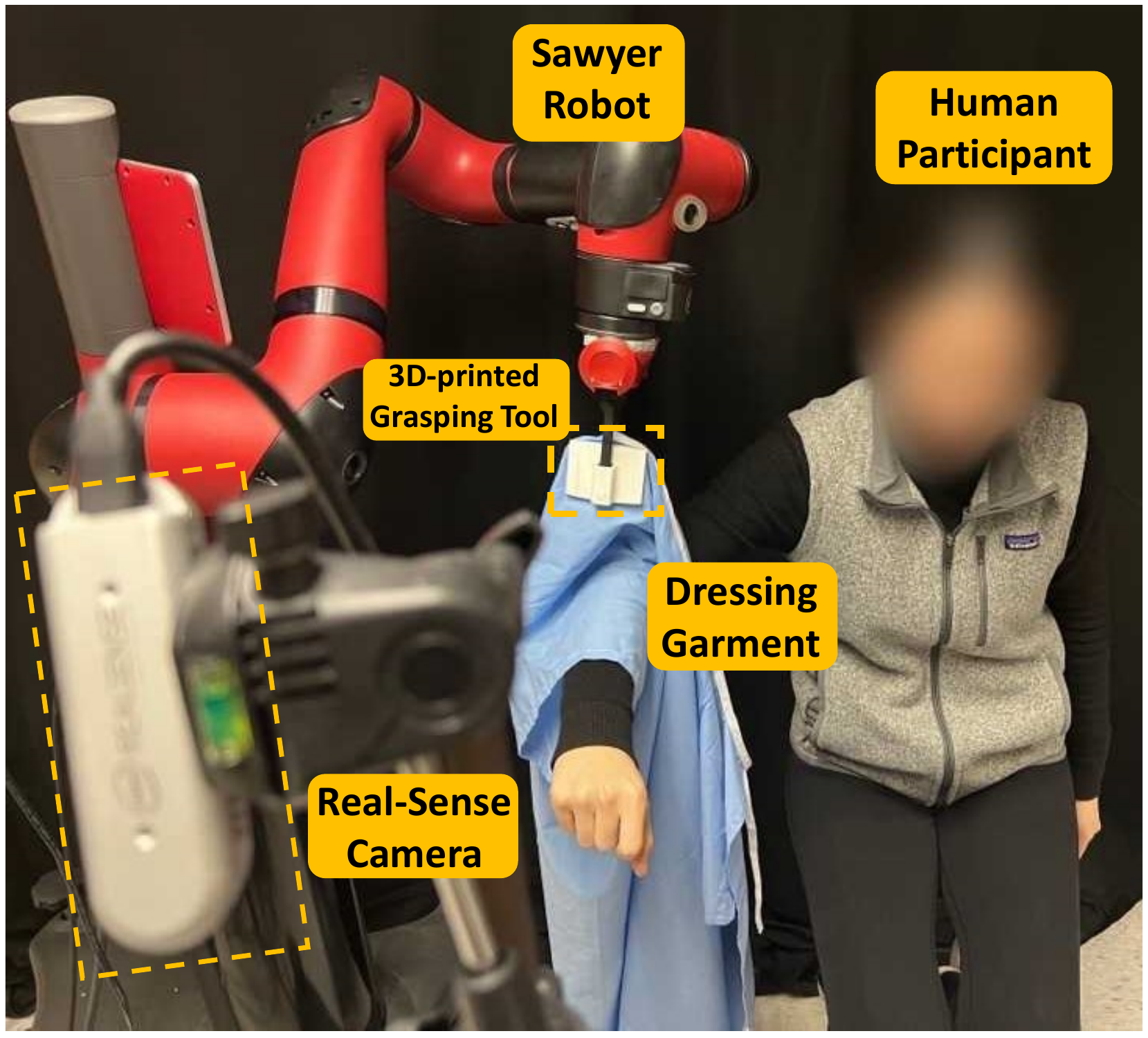}
    \caption{Real-world human study setup.}
    \label{fig:human_study}
    \vspace{-0.1in}
\end{figure}

\begin{figure}[t]
    \centering
    \includegraphics[width=\columnwidth]{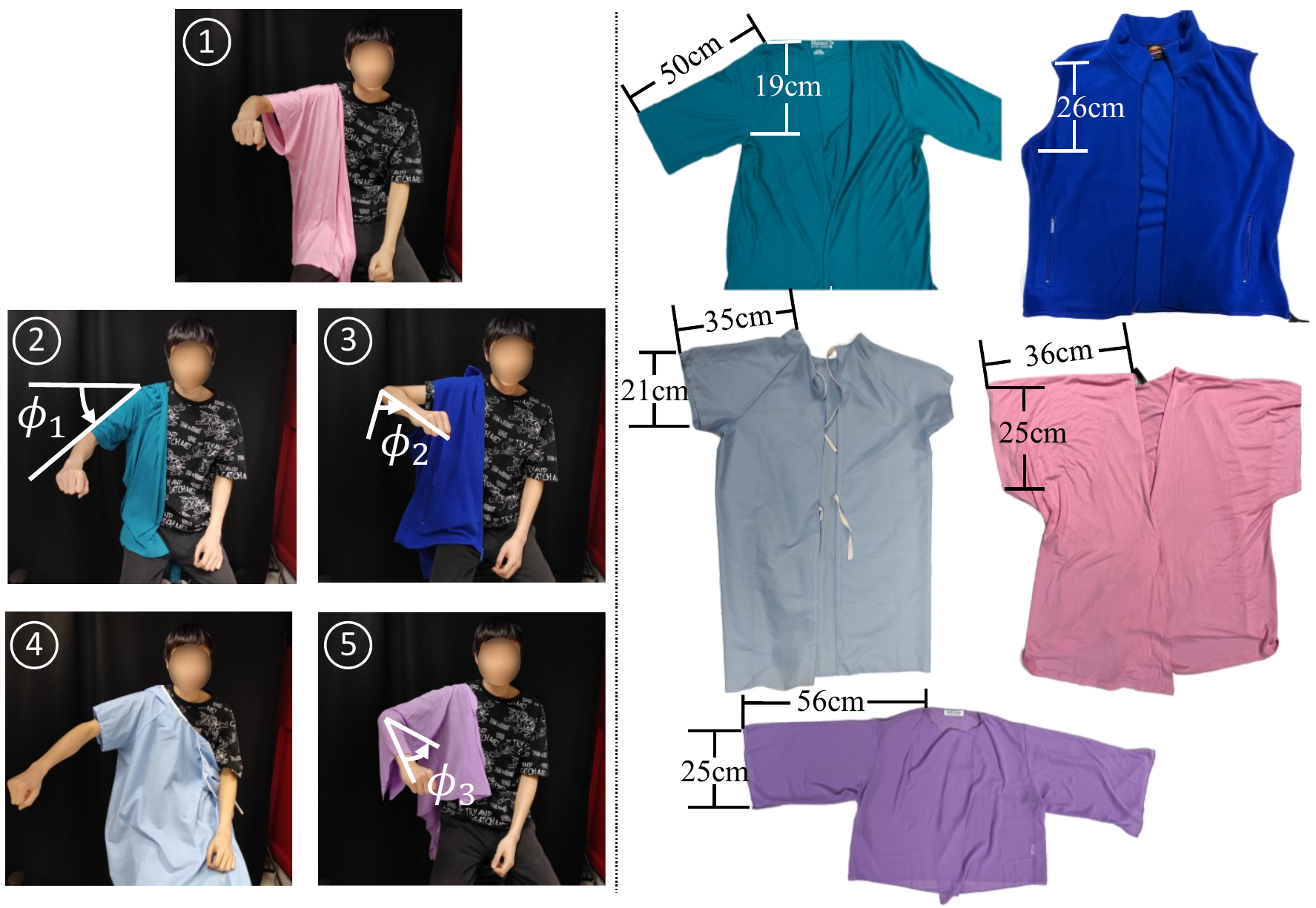}
    \caption{Arm poses and garments used for the human study.}
    \label{fig:pose_and_garment}
    \vspace{-0.2in}
\end{figure}

\section{Real-world Experiments and Human Study}
We perform guided domain randomization (Section~\ref{sec:guided_distillation}) to obtain a distilled policy that is robust for sim2real transfer, and deploy it in the real world, both for dressing a manikin and in a real-world human study. 
Our real-world experiments aim to answer the following questions:
(1) Does our distilled policy outperform the baseline in the real world?
    (2) Does our policy generalize to different people with diverse poses and different garments?

\subsection{Setup}
\begin{figure}
    \centering
    \includegraphics[width=\columnwidth]{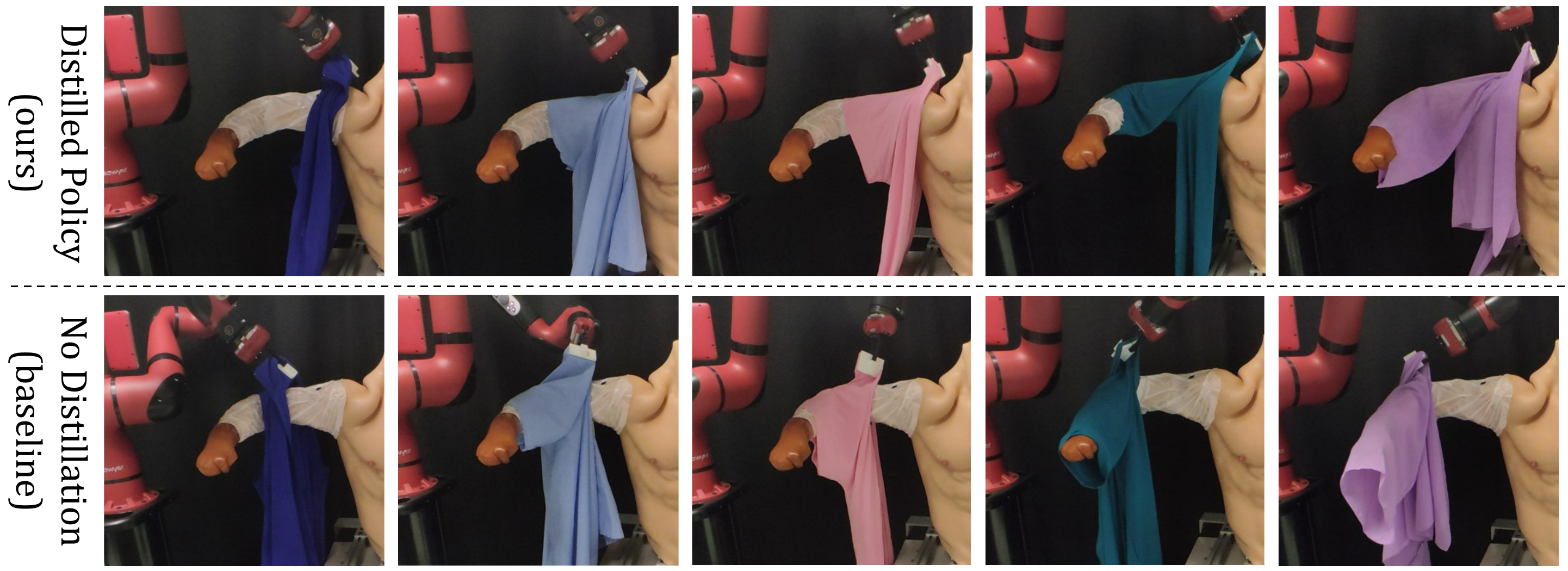}
    \caption{Final state comparison of the Distilled Policy (Ours, top) and No Distillation baseline (bottom) on the manikin.}
    \label{fig:manikin}
    \vspace{-0.1in}
\end{figure}

Fig.~\ref{fig:human_study} shows the real-world setup that we use for the human study, and Fig.~\ref{fig:manikin} shows the manikin setup.
Fig.~\ref{fig:pose_and_garment} shows the arm poses and dressing garments we test. The poses we test in the real world span the pose ranges we train in simulation. We test 5 garments: a sleeveless, non-elastic vest; a short-sleeve, non-elastic hospital gown; a short-sleeve, elastic pink cardigan; a medium-length sleeve, narrow  opening, elastic green cardigan; and a long-sleeve, non-elastic purple cardigan. Note that the real-world garments are not calibrated to the training garments in simulation; for example, the policy is never trained on a sleeveless vest in simulation. 
We use a Sawyer robot to execute the dressing task and an Intel RealSense D435i camera to capture the depth images. 

We use the following quantitative metrics to measure the performance of a dressing trial.
    \textbf{Dressing Success:} We put a marker on the participant's shoulder, at the position that is $80\%$ of the length up the upper arm. If the marker is covered by the garment at the end of the dressing trial, we consider this dressing trial is a success; otherwise it is not.
    \textbf{Upper Arm Dressed Ratio:} We compute the ratio between the ``dressed upper arm distance" and the length of the participant's upper arm. The ``dressed upper arm distance" is measured as the distance from the elbow to the intersection point of the garment and the participant's upper arm (see Fig.~\ref{fig:reward}).
    \textbf{Whole Arm Dressed Ratio:} We compute the ratio  between the ``dressed whole arm distance" (the upper arm dressed distance + the forearm dressed distance) and the length of the participant's whole arm (upper arm length + forearm length). 

\begin{table}[t]
\centering
\scriptsize
\begin{tabular}{c|c|c|c|c|c|c}
\toprule
               & 
               \centering
               \includegraphics[width=0.05\columnwidth]{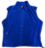}
               & 
               \includegraphics[width=0.04\columnwidth]{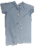}
               & 
               \includegraphics[width=0.05\columnwidth]{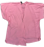}
               & 
               \includegraphics[width=0.07\columnwidth]{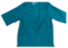}
               &
               \includegraphics[width=0.07\columnwidth]{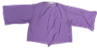}
               & {\begin{tabular}[c]{@{}c@{}} Total\end{tabular}}\\ \hline
{\begin{tabular}[c]{@{}c@{}}Distilled \\ Policy (Ours)\end{tabular}} & 10/10 & 10/10 & 10/10 &  10/10 & 10/10 & 50/50 \\ \hline
{\begin{tabular}[c]{@{}c@{}}No \\ Distillation\end{tabular}} & 0/10   & 0/10  & 0/10 & 0/10 & 0/10 & 0/50     \\ 
\bottomrule
\end{tabular}
\caption{Dressing success rate of the Distilled and No Distillation policies on 5 garments on the manikin.
}
\label{tab:manikin_performance}
    \vspace{-0.2in}
\end{table}

\begin{figure}
    \centering
    \includegraphics[width=\columnwidth]{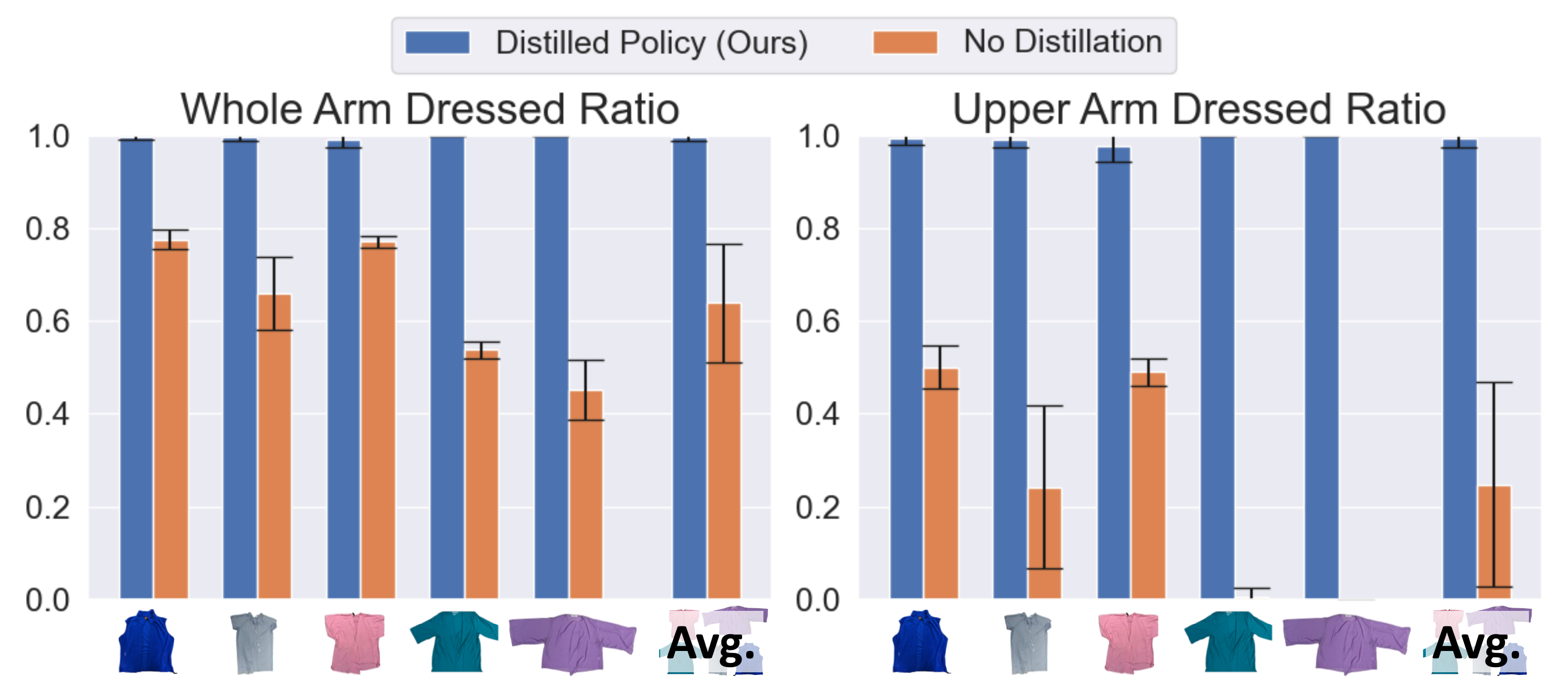}
    \caption{The whole arm (left) and upper arm (right) dressed ratio of the Distilled Policy and the No Distillation baseline on 5 garments on the manikin, averaged across 10 trials for each garment.   
    }
    \label{fig:manikin_performance}
    \vspace{-0.15in}
\end{figure}

\subsection{Comparison on a Manikin}
In order to perform a strictly controlled comparison of the Distilled Policy and the No Distillation baseline, we first conduct experiments on a manikin shown in Fig.~\ref{fig:manikin}.
The manikin has a fixed pose similar to pose 1 in Fig.~\ref{fig:pose_and_garment}.
We conduct 10 dressing trials per garment (across 5 garments), resulting in 50 trials for each method. The numerical results are shown in Table~\ref{tab:manikin_performance} and Fig.~\ref{fig:manikin_performance}. As shown, the Distilled Policy performs vastly better than the No Distillation baseline on all metrics, which resembles performance in simulation. The Distilled Policy successfully covered the marker on the shoulder every trial for all the garments.  Fig.~\ref{fig:manikin} shows the final state achieved by the Distilled Policy and the No Distillation baseline.  For most of the trials, the No Distillation policy can only dress the forearm, and it has trouble correctly turning at the elbow to dress the upper arm.

\begin{figure}
    \centering
    \includegraphics[width=\columnwidth]{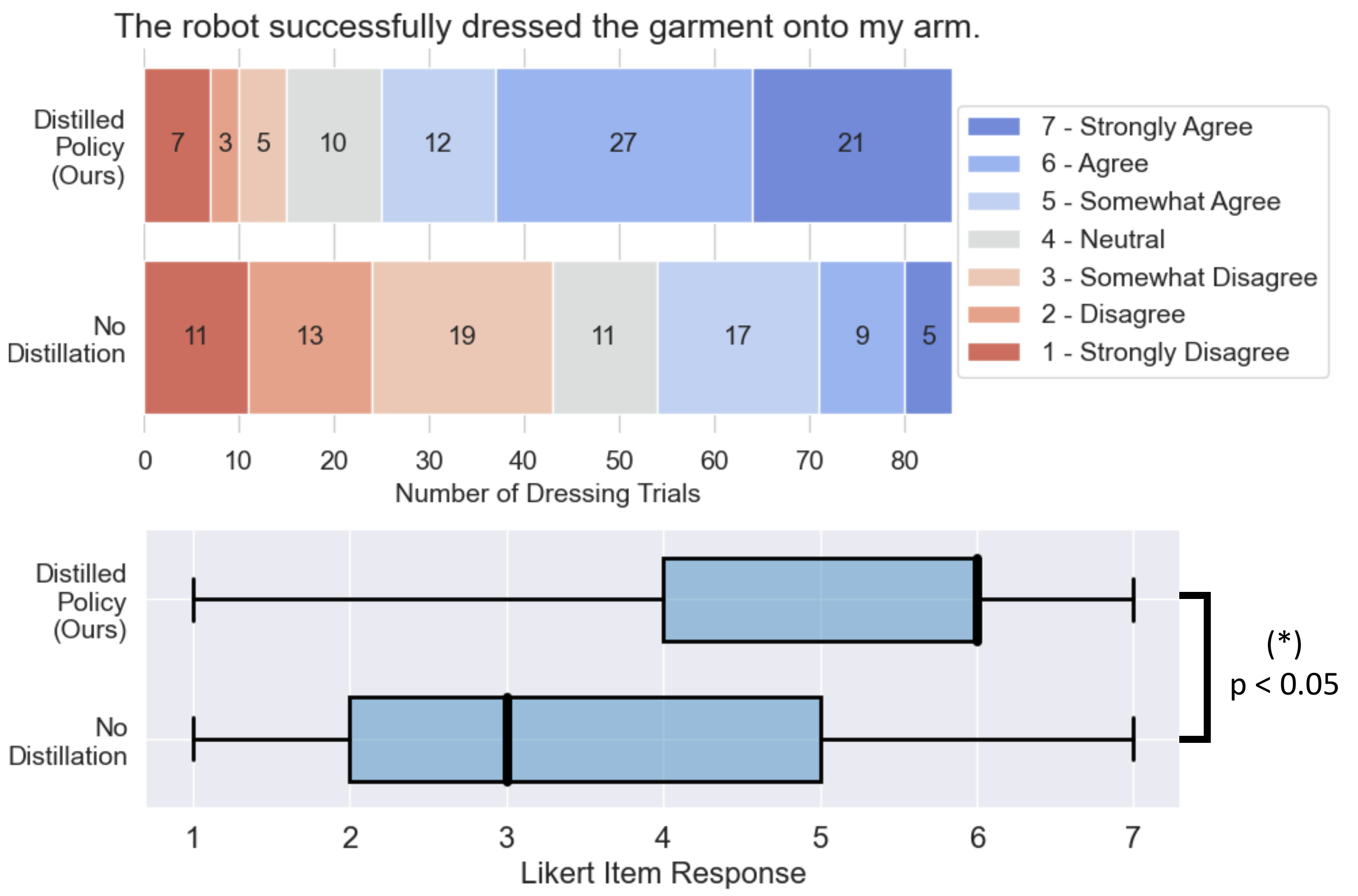}
    \vspace{-0.2in}
    \caption{Likert item responses of the Distilled Policy and the No Distillation baseline, on the 85 dressing trials for which both methods are evaluated on the same poses and garments, shown as a full distribution (top) or box plot (bottom).
    We perform the Wilcoxon signed rank test on the Likert item response and find a statistically significant difference $(p<0.05)$ between our distilled policy and the baseline.
    }
    \label{fig:likert_score_distribution}
    \vspace{-0.1in}
\end{figure}

\begin{table}[t]
\centering
\scriptsize
\begin{tabular}{c|c|c|c}
\toprule
               & \multicolumn{1}{c|}{\begin{tabular}[c]{@{}c@{}}Whole Arm \\ Dressed Ratio\end{tabular}} & \multicolumn{1}{c|}{\begin{tabular}[c]{@{}c@{}}Upper Arm \\ Dressed Ratio\end{tabular}} & \multicolumn{1}{c}{\begin{tabular}[c]{@{}c@{}}Success \\ Rate\end{tabular}} 
               \\ \hline
{\begin{tabular}[c]{@{}c@{}}Distilled \\ Policy (Ours)\end{tabular}} & 
$\mathbf{0.86 \pm 0.19}$  & $\mathbf{0.74 \pm 0.35}$  &  $\mathbf{0.66 \pm 0.47}$ 
\\ \hline
{\begin{tabular}[c]{@{}c@{}}No \\ Distillation\end{tabular}} & 
$0.63 \pm 0.15$ & $0.26 \pm 0.27$ & $0.01 \pm 0.11$ 
\\ 
\bottomrule
\end{tabular}
\caption{Comparison of the Distilled Policy and the No Distillation baseline in a real-world human study. Results are averaged over 17 participants, on the 85 trials for which both methods are evaluated on the same poses and garments.
}
\label{tab:human_study_baseline_comparison}
    \vspace{-0.1in}
\end{table}

\subsection{Human Study Procedure}
\label{sec:human_study_procedure}
We recruit 17 participants in the human study, including 6 females and 11 males.
The age of the participants ranges from 19 to 29. 
For each participant, we conduct 6 trials for each of the 5 garments, totaling 30 trials. Among the 6 trials per garment, we perform 5 trials using our distilled policy, corresponding to the 5 human poses shown in Fig.~\ref{fig:pose_and_garment}. 
To compare against the baseline, we perform the remaining trial using the No Distillation baseline on a pose randomly chosen from the 5 poses. 
We test the No Distillation baseline with only one pose for each garment as we find that most participants experience arm soreness after 30 trials; thus it is impractical to test all poses with the baseline policy. Furthermore, since the No Distillation baseline shows poor performance on a static manikin as shown in Table~\ref{tab:manikin_performance}, it is less likely that it would work on real humans with diverse arm shapes and poses.  
We randomize the test order of the garments, the poses, and the policy for each participant. During the study, the participant is not aware of which policy we are testing.

The procedure of each trial is as follows:
We first show the participant the pose they should imitate, and they lift their right arm to maintain the pose. We then capture a point cloud of the participant's right arm. We move the Sawyer's end-effector, which is already holding the garment, to be positioned near the participant's hand. We then capture a point cloud of only the garment using color thresholding. The color-based segmentation of the garment could be replaced by training a garment segmentation network, such as fine-tuning an existing segmentation network with a small amount of data in our setting~\cite{kirillov2023segment}.
The point clouds of the human arm (recorded statically before the trial), the garment (color thresholded at each timestep), and the end-effector position (obtained using forward kinematics of the robot) are concatenated as input to the policy. 
The participant holds their arm steady throughout the trial. 
The trial terminates if any of the following conditions holds true: (1) if a maximum time step of 75 is reached, (2) if the participant's shoulder is covered by the garment, (3) if the policy is not making any progress in 15 consecutive time steps, and (4) if the participant wishes to have trial stopped. After the trial terminates, we measure dressing distances to compute our evaluation metrics. Finally, at the end of each trial, we provide the participant with a single 7-point Likert item statement ``The robot successfully dressed the garment onto my arm'', which ranges from 1=`Strongly Disagree' to 7=`Strongly Agree'. More details on the human study procedure can be found in Appendix Section C.

\begin{table}[t]
\centering
\begin{tabular}{c|c|c}
\toprule
\multicolumn{1}{c|}{\begin{tabular}[c]{@{}c@{}}Whole Arm \\ Dressed Ratio \end{tabular}} & \multicolumn{1}{c|}{\begin{tabular}[c]{@{}c@{}}Upper Arm \\ Dressed Ratio \end{tabular}} & 
\multicolumn{1}{c}{\begin{tabular}[c]{@{}c@{}}Success \\ Rate\end{tabular}} 
\\ \hline           
 $0.86 \pm 0.17$   & $0.71 \pm 0.34$ & $0.57 \pm 0.49$ \\
 \bottomrule
\end{tabular}
\caption{Performance of our distilled policy, averaged over 17 participants and all 425 dressing trials (not just the 85 trial subset in Table~\ref{tab:human_study_baseline_comparison}).}
\label{tab:human_study_performance}
    \vspace{-0.1in}
\end{table}

\begin{figure}
    \centering
    \includegraphics[width=\columnwidth]{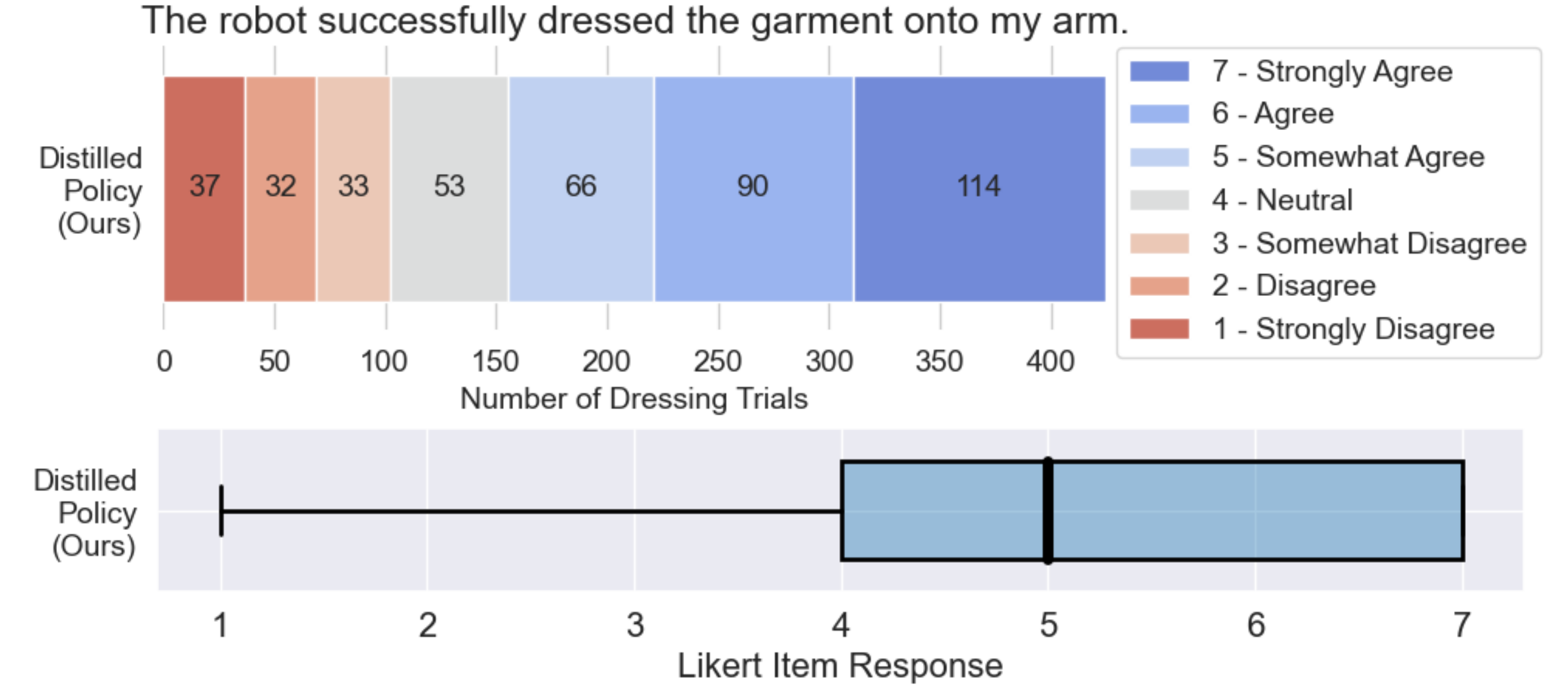}
    \vspace{-0.2in}
    \caption{(Top) Full distribution and (Bottom) box plot of the Likert item responses of the Distilled Policy on
    all 425 dressing trials (not just the 85 trial subset shown in Fig.~\ref{fig:likert_score_distribution}). 
    }
\label{fig:human_study_all_score_box}
    \vspace{-0.1in}
\end{figure}

\begin{figure}[t]
    \centering
    \includegraphics[width=\columnwidth]{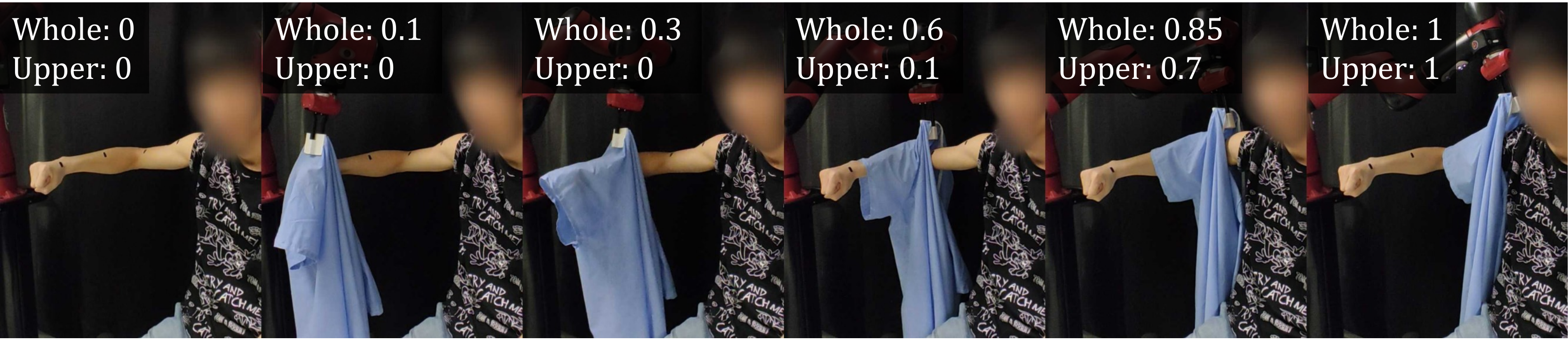}
    \caption{An illustration of the dressed ratios during a  trial.
    }
    \label{fig:illustration_of_dressed_ratio}
    \vspace{-0.1in}
\end{figure}

\begin{figure}
    \centering
    \includegraphics[width=\columnwidth]{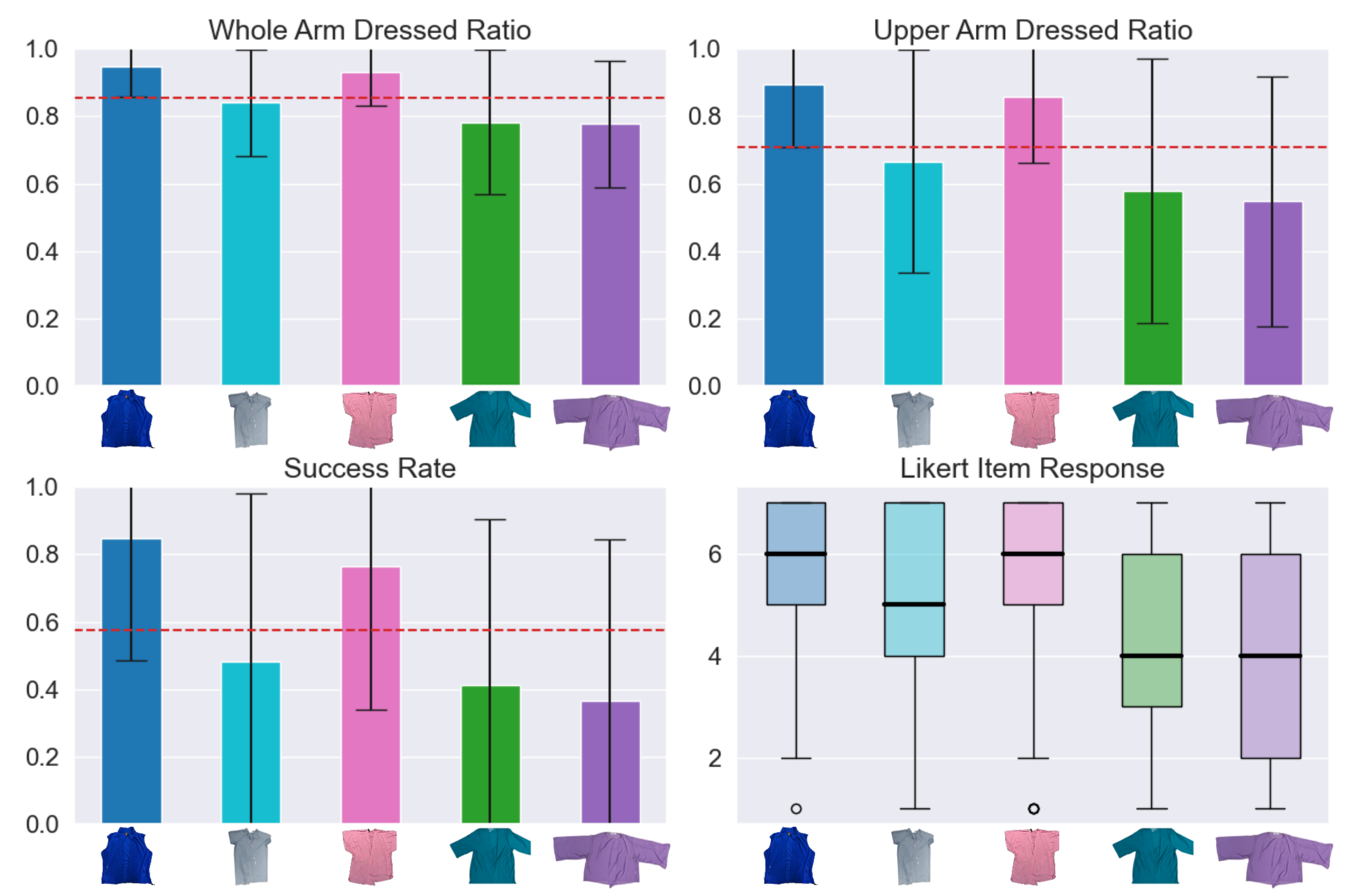}
    \caption{Performance of our distilled policy on different garments. The dashed red line shows the average over all garments.
    }
    \label{fig:garment_performance}
    \vspace{-0.3in}
\end{figure}

\begin{figure}[t]
    \centering
    \includegraphics[width=.8\columnwidth]{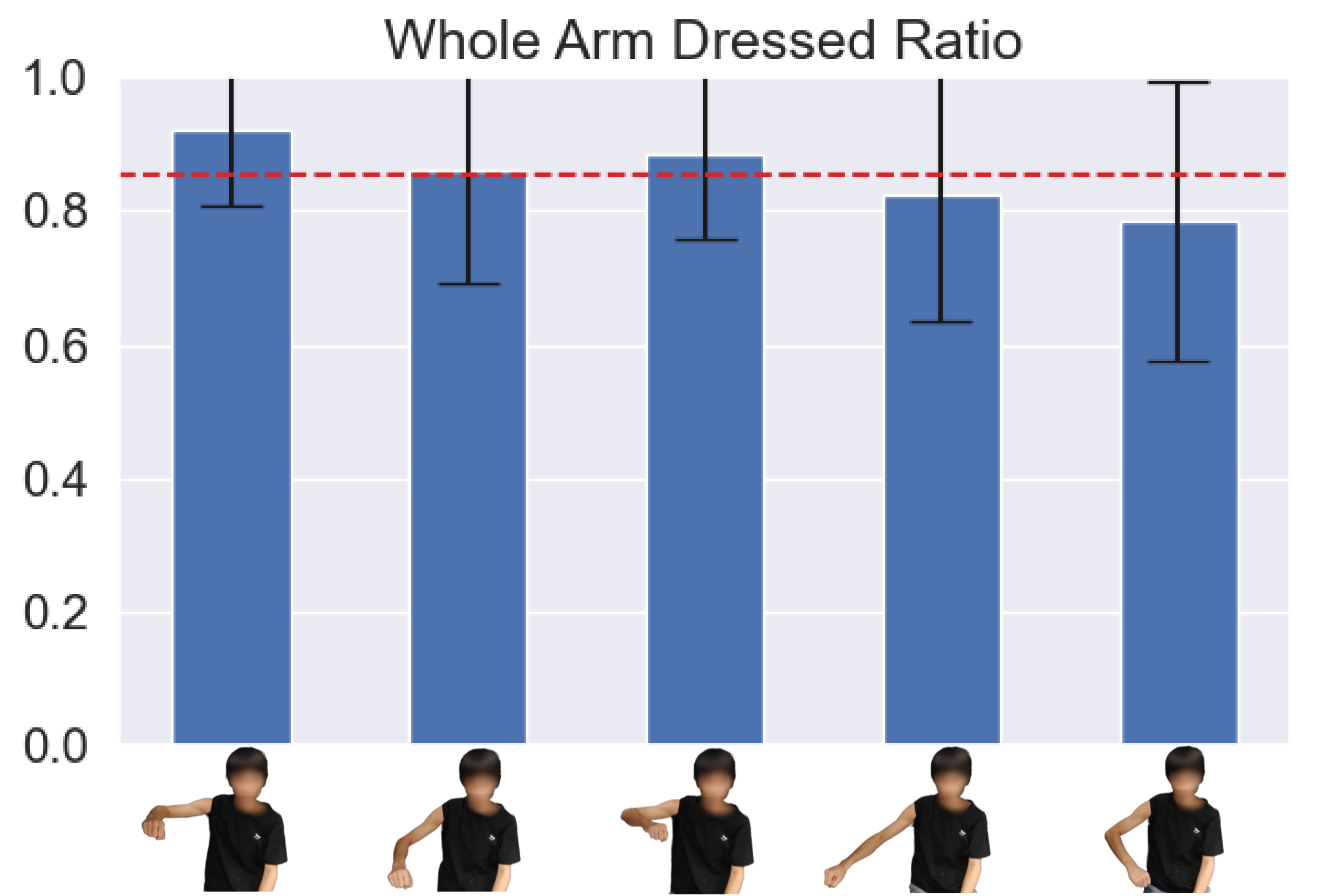}
    \caption{Whole arm dressed ratio of our distilled policy on different poses. The dashed red line represents the average over all poses. Other metrics can be found in Appendix Section C. 
    }
    \label{fig:pose_performance}
    \vspace{-0.3in}
\end{figure}

\subsection{Human Study Results and Analysis}
We first compare the performance of the Distilled Policy against the No Distillation baseline in Table~\ref{tab:human_study_baseline_comparison}. 
The comparison is made on the 85 dressing trials for which both methods are evaluated on the same poses and garments.
As shown, the Distilled Policy outperforms the No Distillation baseline by a large margin under all metrics. 
Fig.~\ref{fig:likert_score_distribution} shows the Likert item response distributions of our distilled policy versus the baseline. Overall, participants agreed at higher rates that the distilled policy successfully dressed the garment, as compared to the baseline. On average,
our distilled policy achieves a median response of 6.0, meaning that the participants ``Agree'' that the robot successfully dressed the garment onto their arm, while the No Distillation baseline achieves a median response of 3.0, meaning the participants ``Somewhat disagree.''  Let $d$ be the difference between the Distilled Policy's Likert item response and the No Distillation baseline's Likert item response. We perform the Wilcoxon signed rank test to test if 
the distribution of $d$ is stochastically greater than a distribution symmetric about zero. We obtain a p-value of $p=0.03125$; hence we find a statistically significant difference $(p<0.05)$ between our distilled policy and the baseline, and the median of the difference is positive.

We now analyze our policy's performance over all 425 dressing trials from the 17 participants. 
Table~\ref{tab:human_study_performance} shows the Distilled Policy's performance under all metrics. 
On average, our policy is able to dress 86\% of the participant's whole arm, and 71\% of the participant's upper arm, achieving a final dressing state similar to the 5th sub-image in Fig.~\ref{fig:illustration_of_dressed_ratio}. 
Fig.~\ref{fig:intro_example} shows snapshots of the dressing trajectories of our policy.
In terms of Likert item responses, as shown in Fig.~\ref{fig:human_study_all_score_box}, our distilled policy achieves a median response of 5.0 (``Somewhat Agree'' that the robot successfully dressed the garment onto the arm). Note that the metrics and Likert item responses are different from those in Table~\ref{tab:human_study_baseline_comparison} and Fig.~\ref{fig:likert_score_distribution} because here the numbers are averaged across all dressing trials, as opposed to the 85 trial subset used for the baseline comparison. 

Fig.~\ref{fig:garment_performance} shows the performance of our method across different garments. It is interesting to see a consistent ordering of the garments under all metrics. Such ordering aligns with human intuitions based on the garment geometry \& materials (see Fig.~\ref{fig:pose_and_garment}): the vest is sleeveless and thus the easiest to perceive and dress (even though the sleeveless vest has never been trained in simulation). The green and purple cardigans, however, have longer sleeves and are harder in terms of both perception and dressing. Meanwhile, although the pink cardigan and the gown have similar sleeve length, the material of the pink cardigan is much more elastic than the hospital gown. This provides some allowance in the end-effector trajectories during the dressing process, leading to better performance. This also holds true for the purple versus green cardigan -- although the opening of the purple cardigan is larger, it is less elastic, which results in the garment getting caught on the participants' arm more frequently, leading to a failure. 

Finally, we analyze the performance of our distilled policy on different poses in Fig.~\ref{fig:pose_performance}. 
We notice that some poses are harder than others. 
For example, the fifth pose with sharp bends at the shoulder and elbow has the lowest performance. 
On the other hand, the first pose, which does not have a sharp bend at either shoulder nor elbow, achieves the highest performance. 
Such a difference is likely due to the fact that sharp bends require more complex end-effector trajectories, and the garment is also more likely to get caught at sharp bends. 
We also note that although the first pose in the human study is similar to the pose of the manikin in Fig.~\ref{fig:manikin}, the success rate on this pose in the human study is lower than the one achieved on the manikin shown in Table~\ref{tab:manikin_performance}. We speculate that such a difference could be due to the diversity of the arm shape and sizes in the human study, as some participants' arm sizes and shapes might be out of our training distribution, resulting in lower performance. Another reason could be that compared to the manikin, real people unconsciously move their arms during the dressing trial, which violates our assumption that the human holds a static arm pose.

\subsection{Failure Cases}
Visual examples of failure cases can be found in our project webpage and also in Appendix Section C.
The first failure case is that the policy gets stuck and cannot output actions that make further progress for the task, e.g., the policy oscillates between moving up and down and does not move forward. This might be due to that the arm pose of the participant is out of the training distribution, or other aspects of the observation cause a sim2real gap. 
Another failure case is the cloth getting caught on the arm. This usually happens when the participant unintentionally moves their arm too much in the dressing process, the policy actions move the gripper too high above the participant's arm, or the policy actions turn too early at the elbow. Due to limited fidelity of the simulator, the garment in simulation is more elastic and can still be dressed even if the robot is pulling high above the arm. However, the garments we test in the real world are less elastic than the simulated ones and they experience greater friction and get caught more easily. We believe that fine-tuning the policy in the real world can help address both failure cases. Since it is difficult to visually detect if a garment is undergoing high friction or has gotten caught on the body, we believe incorporating force-torque sensing can help alleviate these issues. We leave both as future work. 

\section{Out-of-distribution evaluation and generalization of the system}
We conducted a preliminary out of distribution evaluation of our system by relaxing the static arm assumption, i.e., the participants can move their arm during the dressing process. With experiments in both simulation and real world, we observe our system to be robust to small arm movements. 
We first evaluate how our system performs if the participants change their shoulder or elbow joint angles after we capture the initial arm point cloud.
The simulation experiments are conducted on 1 pose sub-region, and the results show that our system is robust to 8.6 degrees of change in shoulder and elbow joint angles (averaged across 3 types of joint angle changes and 5 garments) while maintaining 75\% of the original performance. Please refer to Appendix Section D for detailed experimental results. We also conducted 4 real-world dressing trials with one participant changing their joint angles, including lowering down the shoulder joint for 5 degrees, lowering down the elbow joint for 5 degrees, bending the elbow joint inwards for 5 degrees, and bending the elbow joint inwards for 10 degrees. Our system succeeds in the first 3 trials, and fails for the last trial since the elbow joint angle change is too large.  
We also evaluate our system with one participant performing constant arm motions during the dressing process in the real world. The participant is asked to perform 4 different kinds of arm motions during dressing, including constantly moving their forearm horizontally, constantly moving their forearm in a spherical motion, constantly moving their forearm up and down, and constantly moving their shoulder up and down. The maximum displacement of the arm during the motion was $\pm$10 centimeters. 
Our system succeeds in all 4 kinds of arm motions. Please see our website for videos of these real-world evaluations.

We also evaluate the generalization of the system towards dual-arm dressing.
With the same single-arm dressing assumptions (static arm pose and robot already grasping the garment), our system generalizes to dual-arm dressing. The primary change is to control two robotic arms, one for each sleeve of the garment, which our Dense Transformation policy can handle well by extracting actions corresponding to both of the robot gripper points. We verified this in simulation with preliminary experiments where we successfully train policies to perform dual arm dressing of a hospital gown and a cardigan on a fixed pose (see visualizations on our project website).
We note that dressing over a person's head, such as with a t-shirt, is more complex, due to the need for more dexterous trajectories and awareness of safety considerations.  
We leave such an extension to future work. 

\section{Conclusion}
In this work, we develop a robot-assisted dressing system that is able to dress diverse garments on people with diverse poses from partial point cloud observations, based on a learned policy. 
We show that with careful design of the policy architecture, reinforcement learning (RL) can be used to learn effective policies with partial point cloud observations that work well for dressing diverse garments. 
We further leverage policy distillation to combine multiple policies trained on different ranges of human arm poses into a single policy that works over a wide variety of different poses.
We propose guided domain randomization for effective and robust sim2real transfer.
We perform comprehensive real-world evaluations of our system on a manikin, and in a human study with 17 participants of varying body size, poses, and dressed garments.
On average, our system is able to dress 86\% of the length of the participants' whole arm, and 71\% of the length of the participants' upper arm across 425 dressing trials.
    We hope this work will serve as a foundation for future research to develop more robust and effective dressing systems.

\section*{Acknowledgement}
This work was supported by the National Science Foundation under Grant No. IIS-2046491, and NVIDIA Corporation for their academic hardware grant.
Any opinions, findings, and conclusions or recommendations expressed in this material are those of the author(s) and do not necessarily reflect the views of the National Science Foundation, or NVIDIA Corporation.

\IEEEpeerreviewmaketitle

\bibliographystyle{plainnat}
\bibliography{references}
\clearpage
\newpage
\onecolumn

\begin{appendices}

  \begingroup
  \centering
  
    \centering
    {\fontsize{28pt}{28pt}\selectfont Appendix}\\
    \vspace{1em}
    {\fontsize{20pt}{20pt}\selectfont One Policy to Dress Them All: Learning to Dress\\People with Diverse Poses and Garments}
    \vspace{2em}
  
  \endgroup

\newcommand{\rebuttal}[1]{\textcolor{red}{#1}}
\newcommand{\tabincell}[2]{\begin{tabular}{@{}#1@{}}#2\end{tabular}}
\newcommand{\theHalgorithm}{\arabic{algorithm}}
\newtheorem{prop}{Proposition}
\renewcommand{\figurename}{Supplementary Figure}
\renewcommand{\tablename}{Supplementary Table}

\section{System Implementation Details}
\subsection{Full Reward Function}
As mentioned in the main paper Section IV.B, in addition to the main reward $r_m$ that measures the progression of the dressing task, we have three additional reward terms, detailed as follows:
\begin{itemize}
    \item The second reward term is a force penalty $r_f$ that prevents the robot from applying too much force through the garment to the person. Too much force will not only hurt the person in the real world but also lead to unrealistic simulation as the cloth will penetrate through the arm. Specifically, let $f$ be the total applied force from the garment to the human, and $f_{max}$ be the max threshold, this reward is computed as $r_f = -0.001 \cdot  \max(f - f_{max}, 0)$. We set $f_{max}$ to be $1000$ units as measured by the simulator, which is an empirical threshold value when the cloth starts penetrating. We note that due to simulation modeling inaccuracies, this force number given by the simulator does not correspond to 1000 Newtons of force in the real world.   
    \item The third reward term is a contact penalty $r_c$ that prevents the robot end-effector from moving too close to the person. Let $d_e$ be the shortest distance between the end-effector and the arm, this reward is computed as $r_c = -0.01 \cdot \mathbf{1}(d_e < d_{min})$, where $\mathbf{1}(d_e < d_{min}) = 1$ if $d_e < d_{min}$ and 0 otherwise. $d_{min}$ is set to be $1$ cm in our experiments. 
    This reward term, as well as the previous one, are used to keep the dressing process safe and comfortable for the person.
    \item The last reward term is a deviation penalty $r_d$ that discourages the garment center from moving too far away from the arm. Let $d_g$ be the shortest distance from the garment center $p^g_{center}$ to the arm,  
    this reward is given by $r_d = 0.02$ if $d_g < 3$ cm, $r_d = -0.05$ if $d_g > 7.5$ cm, and $r_d = 0$ otherwise. 
\end{itemize}

The full reward function we use is: $r = r_m + r_f + r_c + r_d$.

\subsection{Observation Domain Randomization}
\label{sec:obs_randomization}
We pre-process the point cloud by filtering
it with a voxel grid filter: we overlay a 3d voxel grid over the partial point cloud and then take
the centroid of the points inside each voxel to obtain a voxelized point cloud.
This preprocessing step is done both in simulation training and in the real world, which makes our
method agnostic to the density of the observed point cloud and thus more robust during sim2real transfer. We use a voxel size of $6.25$ cm.

As mentioned in main paper Section IV.D, we perform randomization on the garment point cloud observation for robust sim2real transfer. 
The randomizations we perform in simulation are as follows. Note that all these randomizations are just for the observations; the underlying simulator state or physics (e.g., the garment mesh) is not randomized. 
\begin{itemize}
\item Random cropping of garment point cloud. 
To make the garment point cloud observation more aligned between the simulation and the real world, we perform cropping to remove the bottom part of the garment. After the bottom part are cropped, the remaining part (which mostly contains the sleeve part of the garment) is much more similar between simulation and the real world, reducing the sim2real gap. 
Formally, given the garment point cloud $P^g$ and the human finger point $p^h_{finger}$, we remove garment points that are $\delta_1$ cm lower than the human finger point, i.e., the following points are kept:
\begin{equation}
    \{p_i \in P^g ~|~ p_i[z] > p^h_{finger}[z] - \delta_1\},
\end{equation}
where $p[z]$ is the z coordinate of the point, and $z$ is the gravity axis (\figurename~\ref{fig:coordinate} shows the coordinate system we use in the real world). 
 $\delta_1$ is uniformly randomly sampled from $[10, 25]$ cm in our experiments.

In the real world, we obtain the garment point cloud $P^g$ via color thresholding. Color thresholding will usually return some additional noise points that are not part of the garment, especially when environment conditions such as lighting changes. To account for such error in the real world, we add more cropping to remove the noise points returned by color thresholding. Specifically, all garment points that are $\delta_2$ cm above the robot gripper point $P^r$, or $\delta_3$ cm forward the gripper point (the forward direction is along the x axis shown in \figurename~\ref{fig:coordinate}) are removed. Formally, only points satisfying the following conditions are kept:
\begin{equation}
    \{p_i \in P^g ~|~ p_i[z] < P^r[z] + \delta_2, ~p_i[x] < P^r[x] + \delta_3\},
\end{equation}
where $x$ is the coordinate axis aligning with the forward moving direction of the robot (See \figurename~\ref{fig:coordinate}). In our experiments, we uniformly randomly sample $\delta_2$ from $[1, 5]$ cm, and $\delta_3$ from $[1, 5]$ cm. 
In summary, we perform the cropping to only keep the following garment points:
\begin{equation}
    \{p_i \in P^g ~|~ p_i[z] > p^h_{finger}[z] - \delta_1, ~p_i[z] < P^r[z] + \delta_2, ~p_i[x] < P^r[x] + \delta_3\}
\end{equation}

\begin{figure}
    \centering
    \includegraphics[width=.5\textwidth]{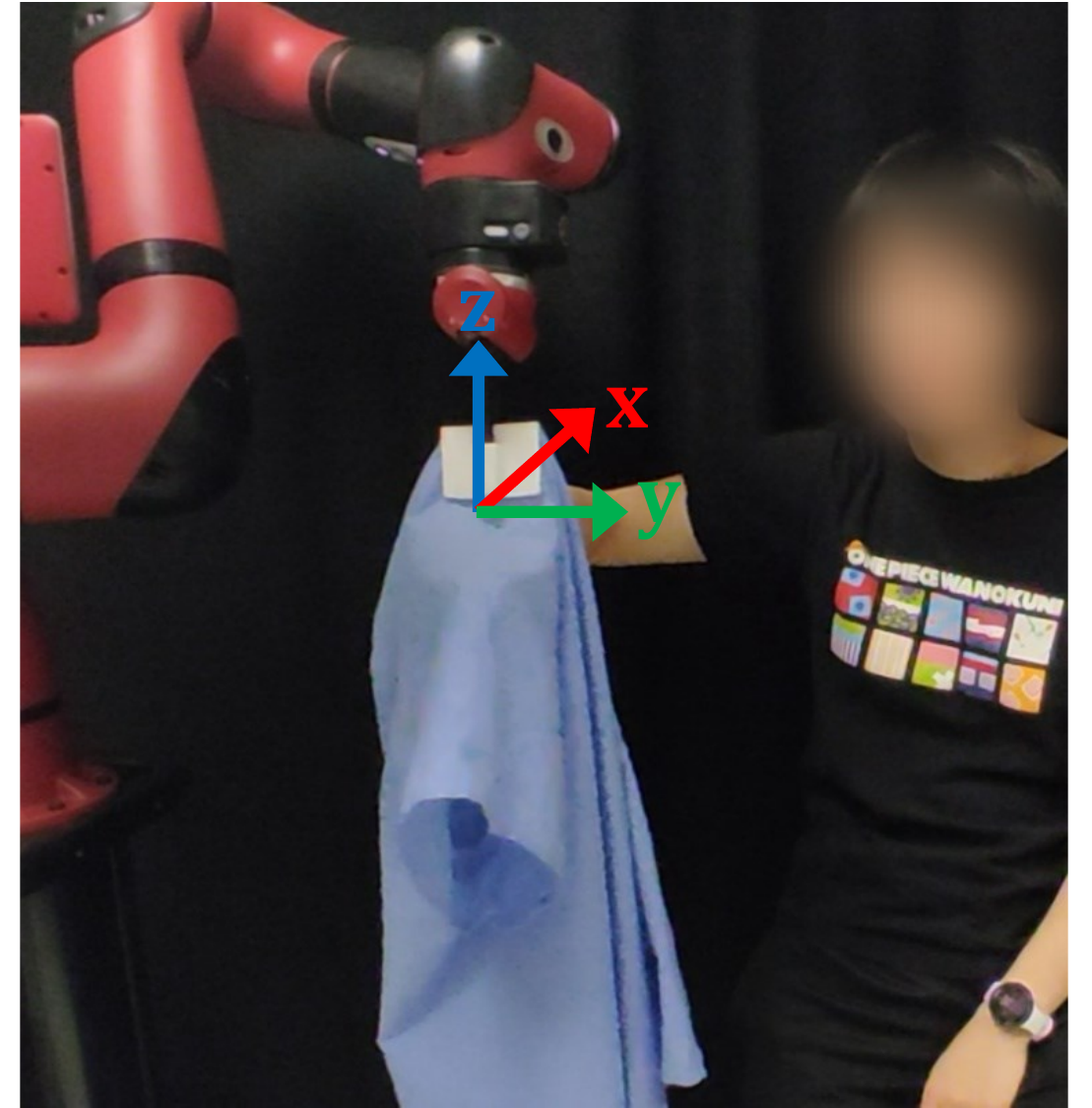}
    \caption{Coordinate system in the real world.}
    \label{fig:coordinate}
\end{figure}

\item Random dropping of garment points.
In the real world, the part of garment points that are grasped by the tool is occluded by the tool (See Fig. 4 in the main paper for an illustration of the 3D-printed grasping tool we use in the real world). To account for this issue, we remove garment points that are near the gripper point in simulation as well. Specifically, any garment points that are $\delta_4$ cm within the gripper point $P^r$ are removed from the observation. Mathematically, points satisfying the following condition are removed:
\begin{equation}
        \{p_i \in P^g ~|~ ||p_i - P^r ||_2 < \delta_4\}
\end{equation}
In each training episode of SAC, we perform this operation of removing points near the gripper with a probability of $0.5$. We set $\delta_4 = 9.375$ cm in our experiment. 

\item Random erosion of the garment.
To account for the geometric differences between the garment in simulation and in the real world, we add random erosion on the garment depth image. The erosion is done on the depth image of the garment before de-projecting it to a point cloud. The kernel size for the erosion is randomly sampled from $[0, 3, 5, 7, 9, 11, 13, 15, 17, 19]$, where $0$ means no erosion will be used.

\item Random dilation of the garment.
To account for the geometric differences between the garment in simulation and in the real world, we add random dilation on the garment depth image. The dilation is done on the depth image of the garment before de-projecting it to a point cloud. The kernel size for the dilation is randomly sampled from $[0, 3, 5, 7, 9, 11, 13, 15, 17, 19]$, where $0$ means no erosion will be used. If the sampled erosion and dilation kernel sizes are both non-zero, we do either erosion or dilation with equal probability, and do not perform them at the same time. 

\item Random noise added to gripper position. 
In the real world, we use a side-view camera and perform camera-to-robot calibration, so we can transform garment and arm point clouds into the robot frame. On the other hand, the robot gripper position is obtained via forward kinematics and is already in the robot frame, thus does not need the calibration and is perfectly accurate in the robot frame. 
Due to camera calibration errors, there will be minor errors when transforming the garment and arm point clouds into the robot frame. To account for such error, we add random noise to the gripper position in simulation. As we use PointNet++ as our policy architecture and it is translation-invariant (i.e., translating the whole input point cloud does not change the output), adding a random noise to the gripper position is equivalent to adding a random noise to all the garment and arm point clouds, thus mimicking the real-world calibration error. Note that the noise is only added to the gripper point in the policy observation, but not the actual gripper in simulation.
At the beginning of each SAC training episode, we uniformly sample a noise from $[-3.125, 3.125]$ cm, and the noise is added to the gripper position throughout this episode.

\end{itemize}

When deploying the policy in the real world, we perform the cropping of the garment point cloud and fix $\delta_1 = 15$ cm, $\delta_2 = 2$ cm, and $\delta_3 = 1$ cm. We do not perform random dropping, erosion, or dilation of the garment points in the real world, and we do not add noise to the gripper position in the real world.

\subsection{Simulation Training Details}
We use PointNet++~\cite{qi2017pointnet++} as both the policy and the Q function network architecture. We use implementation of PointNet++ in Pytorch Geometric~\cite{Fey/Lenssen/2019}.
For the dense transformation policy, we use the segmentation-type PointNet++. We use 2 set abstraction layers followed by a global max pooling layer, 3 feature propagation layers, and followed by a Multi-layer Perceptron (MLP). The set abstraction radius are 0.05 and 0.1, and the sampling ratios are 1 and 1. 
The numbers of nearest neighbors for feature propagation layers are 1, 3, and 3. The final MLP is of size $[128, 128]$.
For the Q function, we use the classification-type PointNet++. We use 2 set abstraction layers, followed by a global max pooling layer, and a MLP. The set abstraction radius are 0.05, 0.1, and the sampling ratios are 1, 1.  The final MLP is of size $[128, 128]$.

We use SAC~\cite{haarnoja2018soft} as the RL training algorithm. The learning rates  for the actor, critic, and the entropy temperature alpha are all $1e-4$. The delayed actor update frequency is 4, i.e., we update the actor once every time the critic is updated 4 times. We use a replay buffer size of 400000, and a batch size of 64. The soft update parameter $\tau$ for the critic target network is set to be $0.01$.  Each episode in simulation has 150 time steps. We use the Adam~\cite{kingma2014adam} optimizer to train both the policy and the Q network.



\begin{figure}
    \centering
    \begin{tabular}{ccc}
           \includegraphics[height=0.15\textwidth]{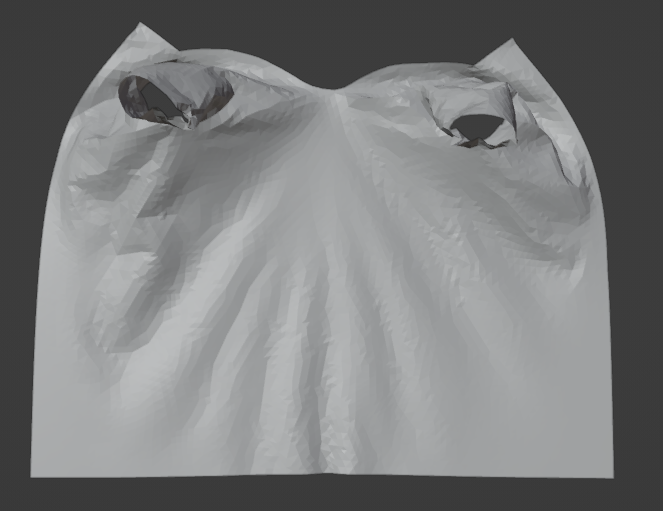}  & 
    \includegraphics[height=0.15\textwidth]{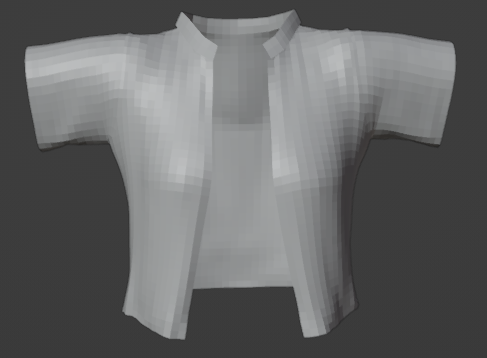}
           &      \includegraphics[height=0.15\textwidth]{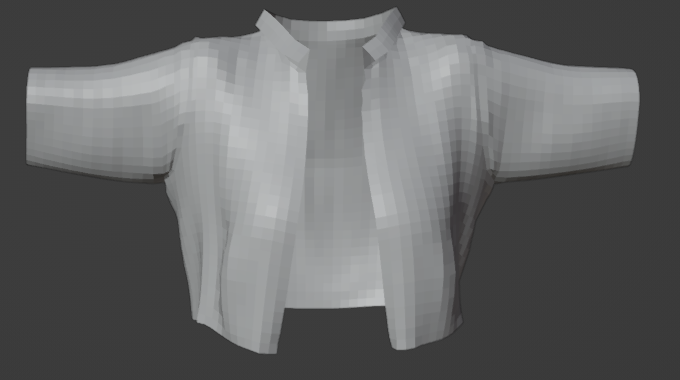}\\
       Garment 1  & Garmen 2  & Garment 3 \\
\includegraphics[height=0.15\textwidth]{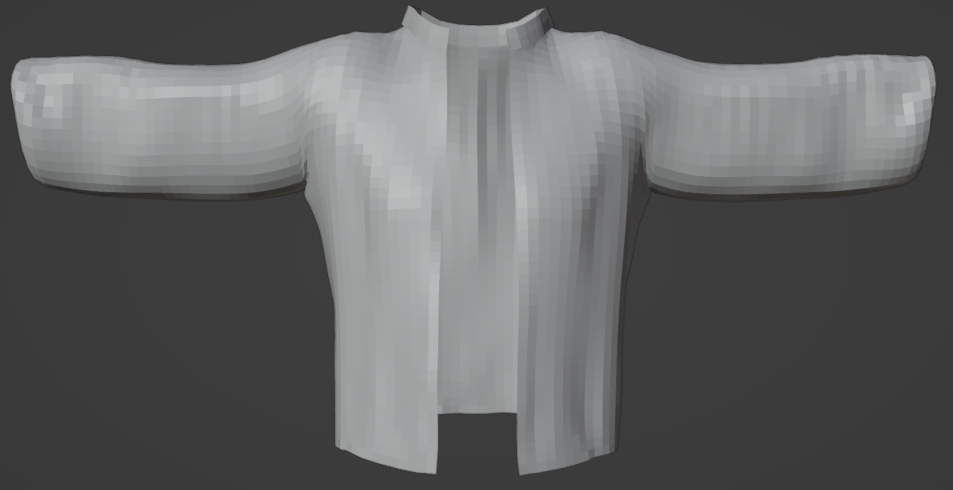} &  \includegraphics[height=0.15\textwidth]{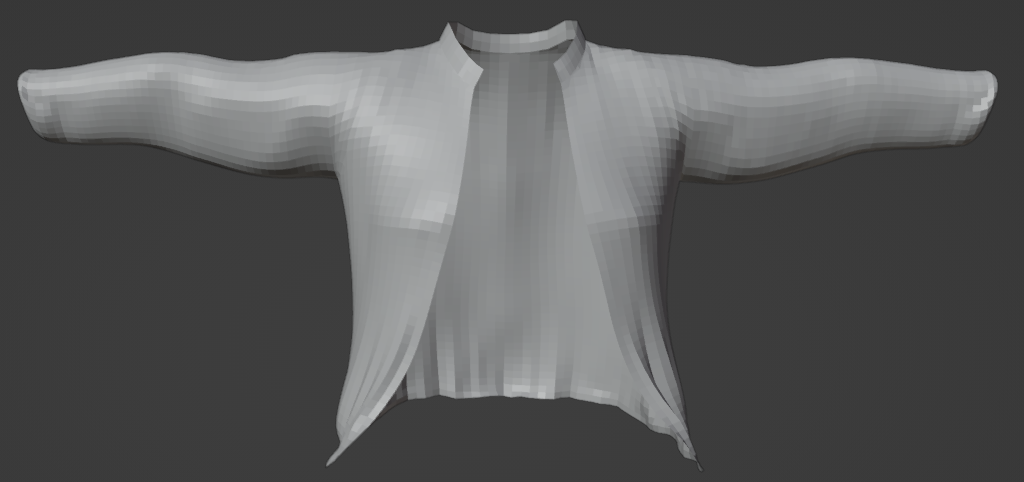} \\
        Garment 4 & Garment 5  
    \end{tabular}
    \caption{Garments we used in simulation training. 
    The first one is a hospital gown, followed by 4 cardigans. The garments are in ascending order in terms of sleeve length. }
    \label{fig:garments}
\end{figure}

\begin{figure}
    \centering
    \includegraphics[width=0.19\textwidth]{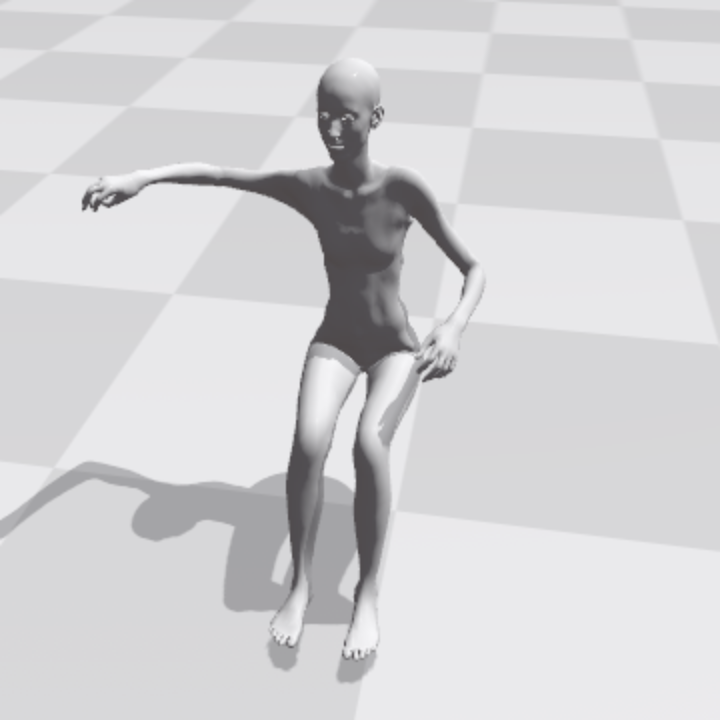}
    \includegraphics[width=0.19\textwidth]{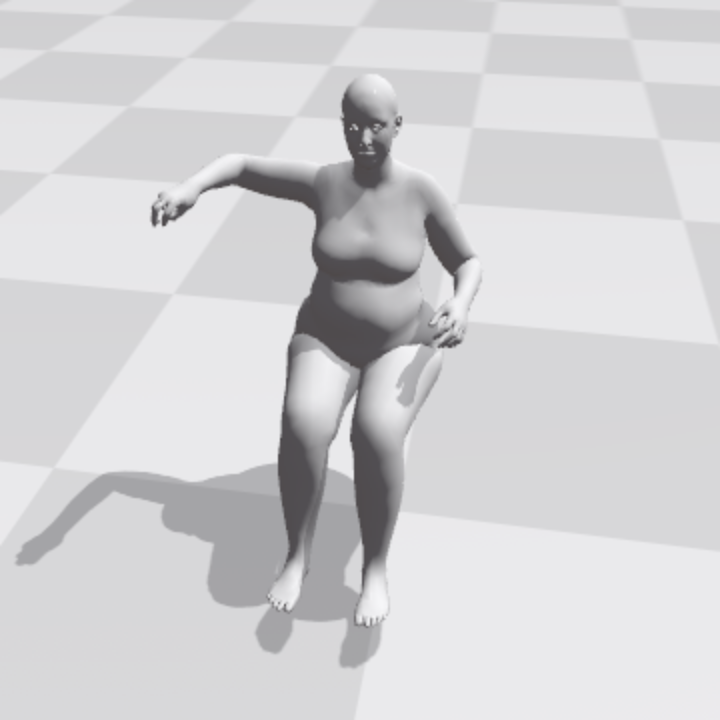}
    \includegraphics[width=0.19\textwidth]{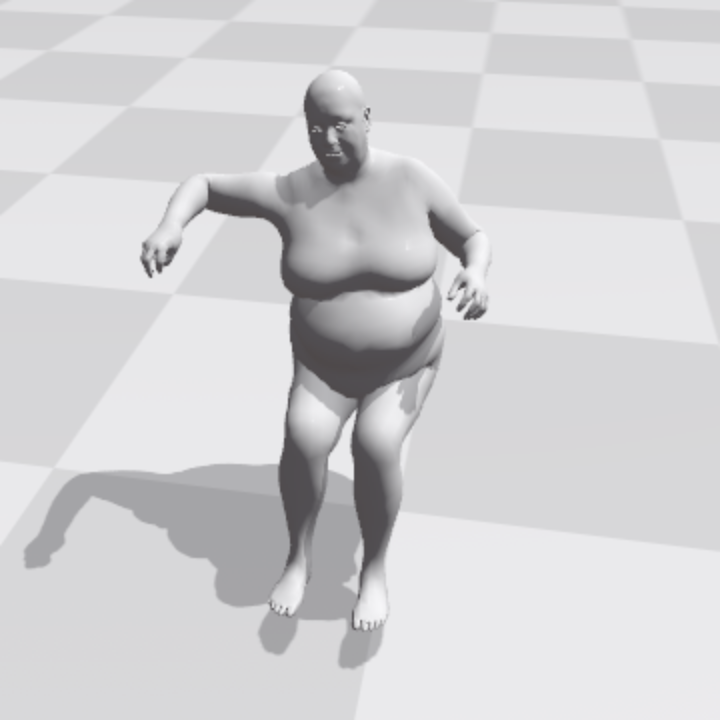}
    \includegraphics[width=0.19\textwidth]{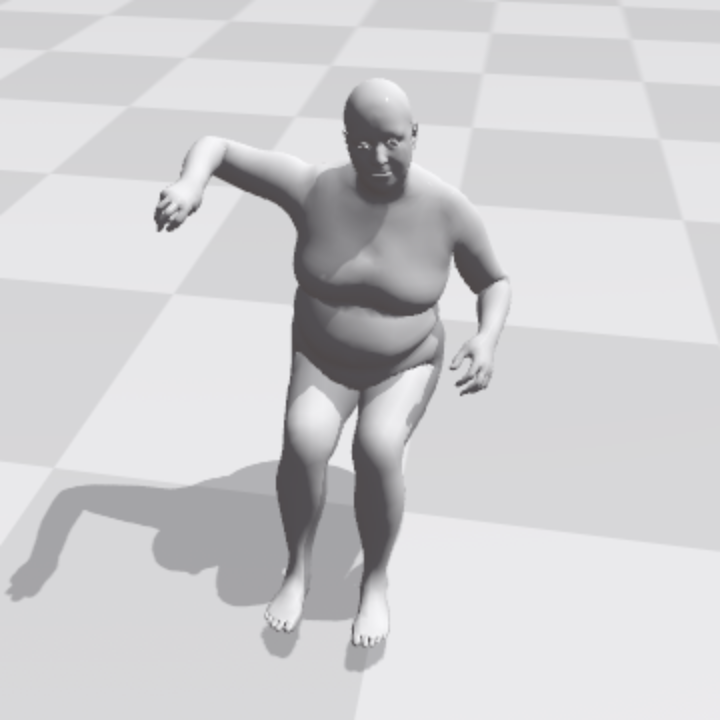}
    \includegraphics[width=0.19\textwidth]{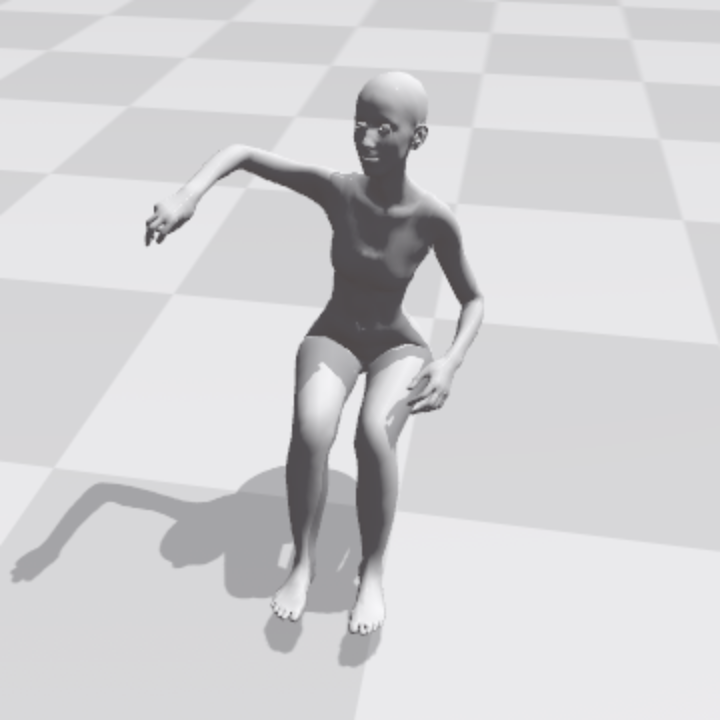}
    \includegraphics[width=0.19\textwidth]{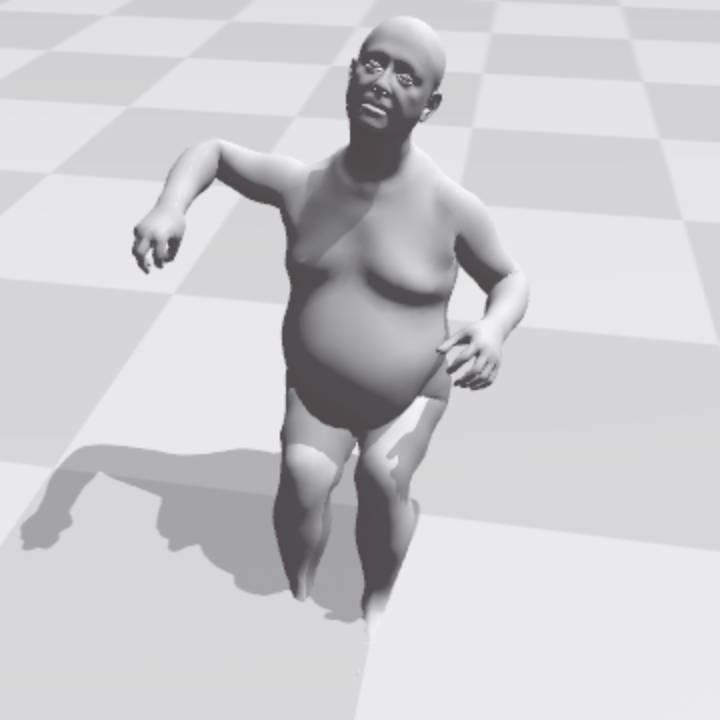}
    \includegraphics[width=0.19\textwidth]{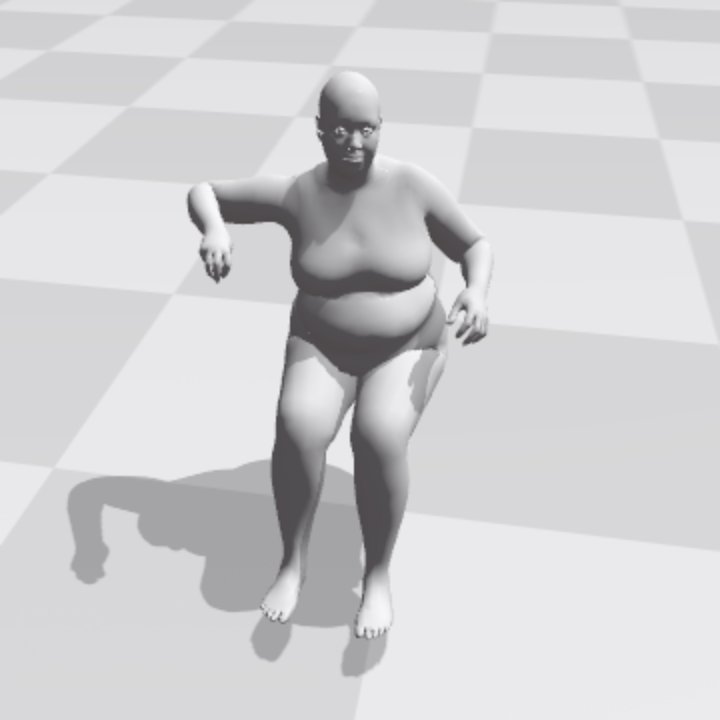}
    \includegraphics[width=0.19\textwidth]{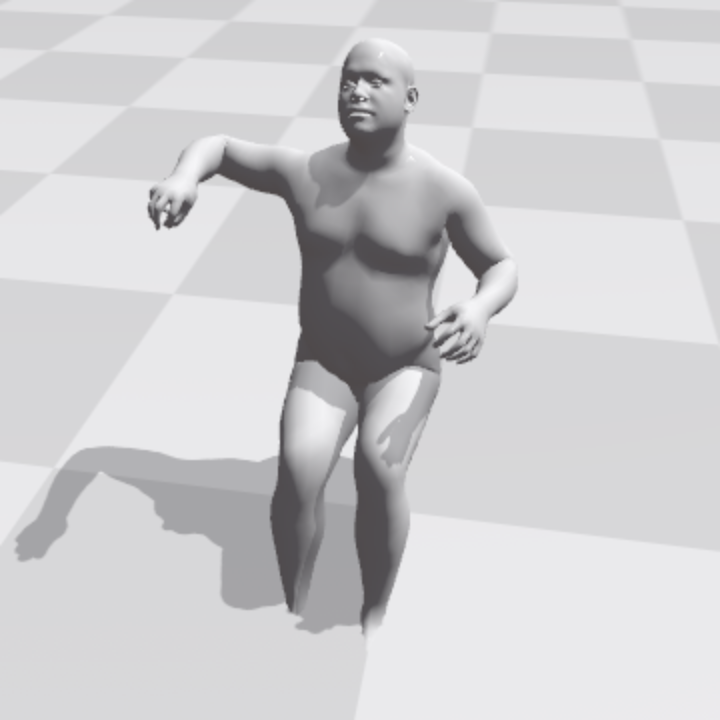}
    \includegraphics[width=0.19\textwidth]{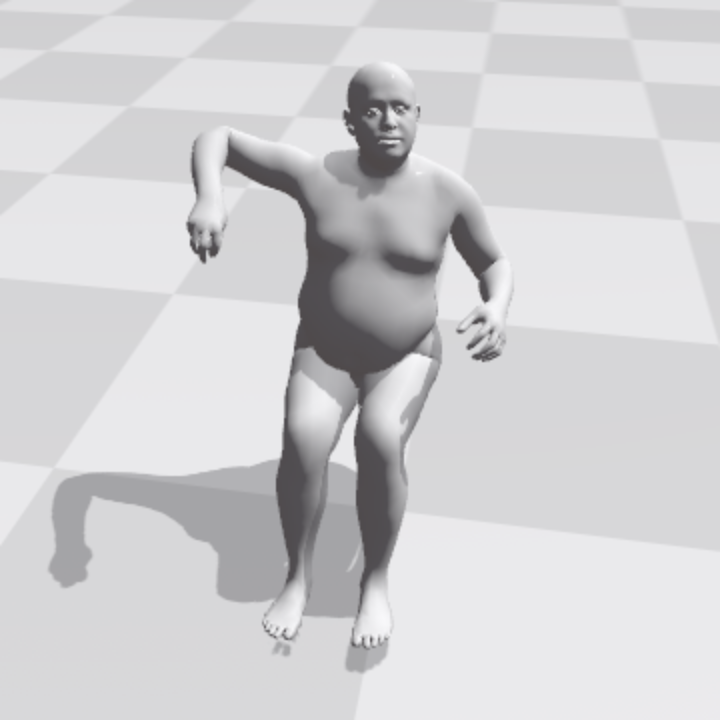}
    \includegraphics[width=0.19\textwidth]{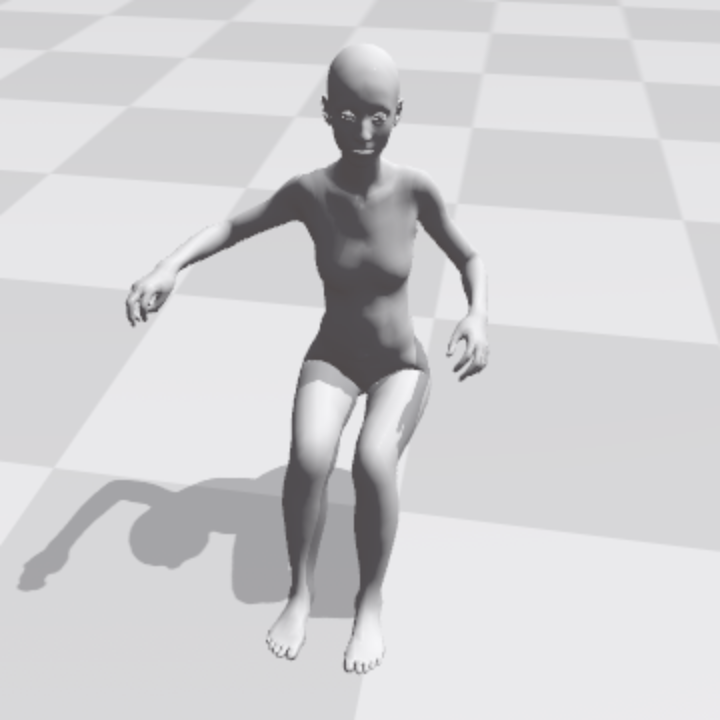}
    \includegraphics[width=0.19\textwidth]{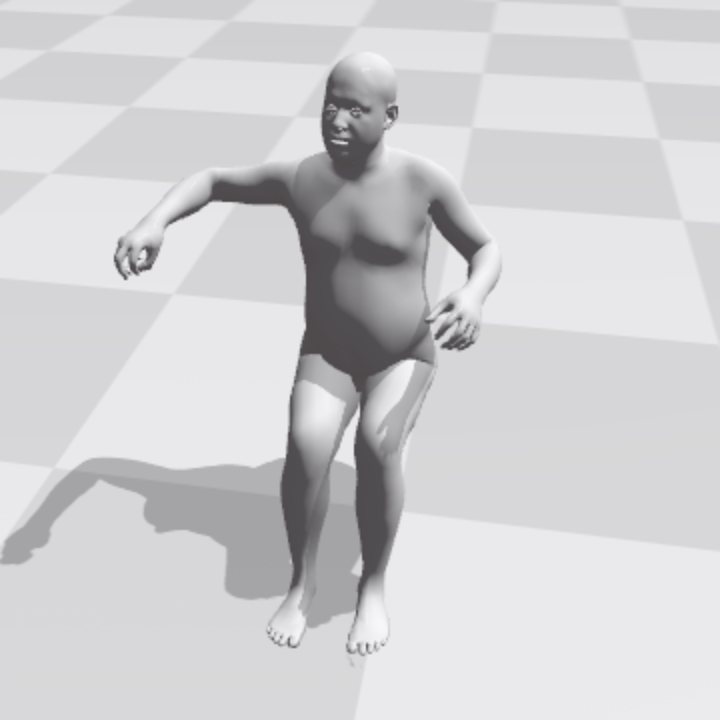}
    \includegraphics[width=0.19\textwidth]{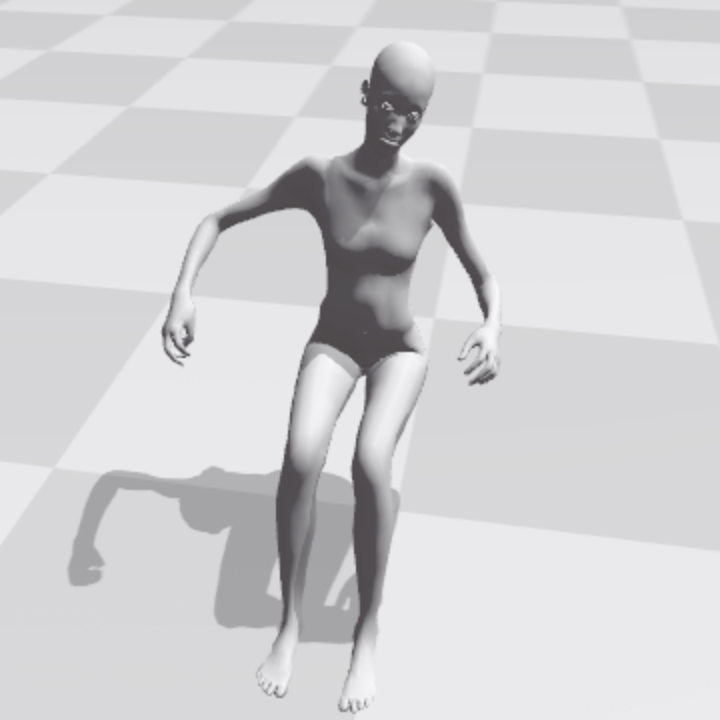}
    \includegraphics[width=0.19\textwidth]{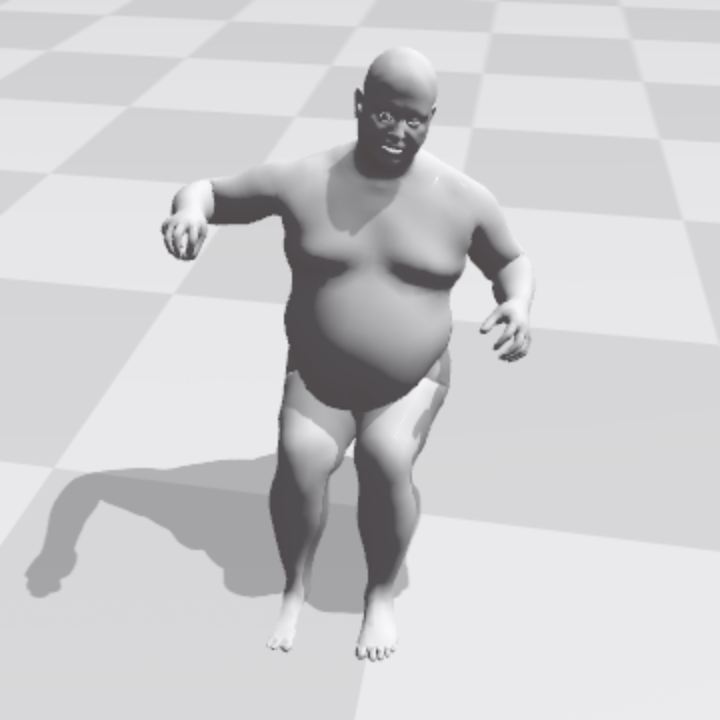}
    \includegraphics[width=0.19\textwidth]{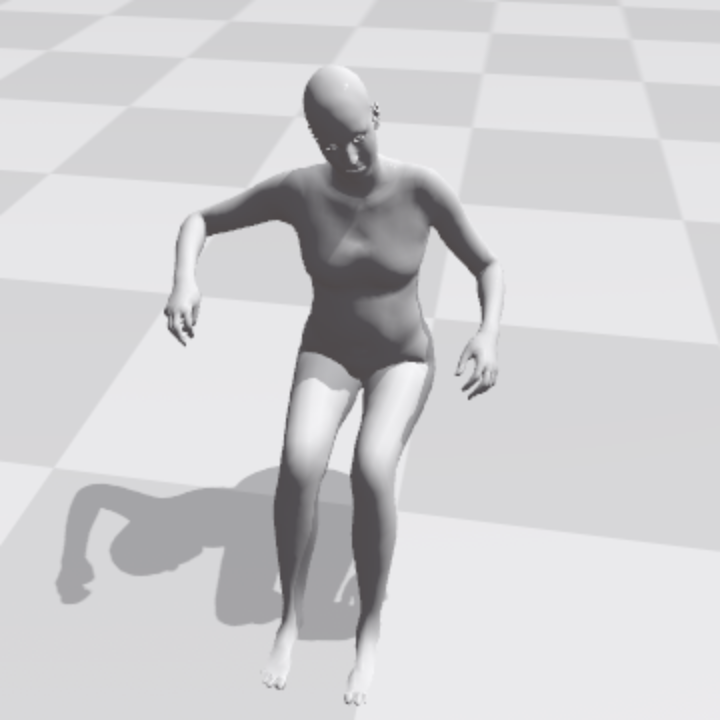}
    \includegraphics[width=0.19\textwidth]{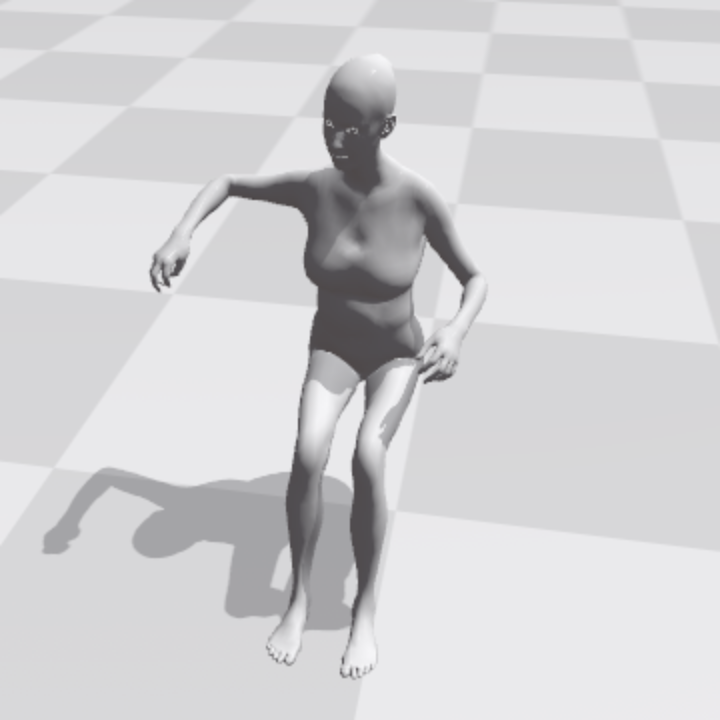}
    \includegraphics[width=0.19\textwidth]{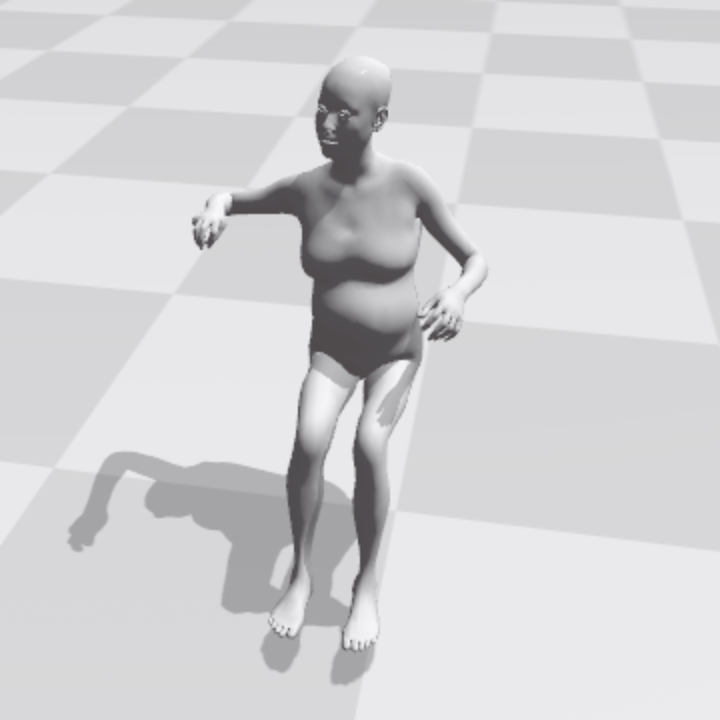}
    \includegraphics[width=0.19\textwidth]{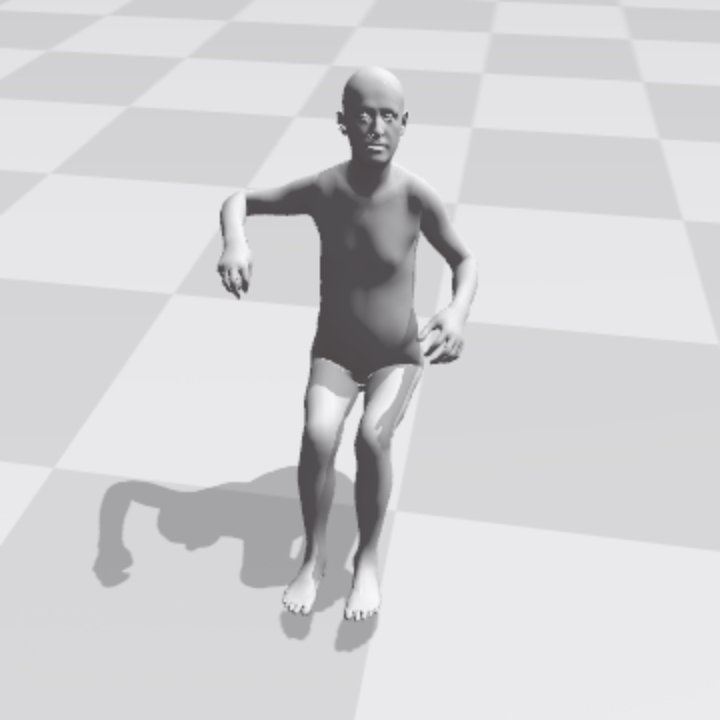}
    \includegraphics[width=0.19\textwidth]{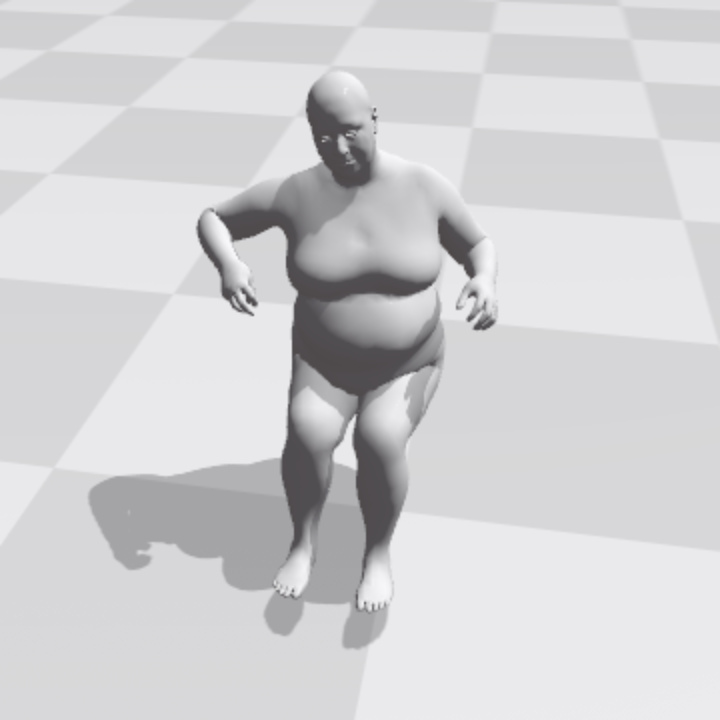}
    \includegraphics[width=0.19\textwidth]{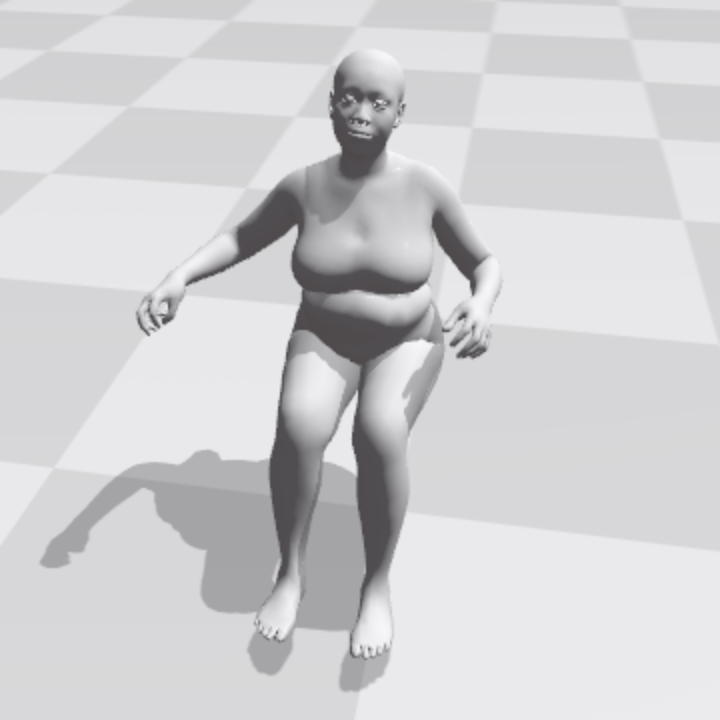}
    \includegraphics[width=0.19\textwidth]{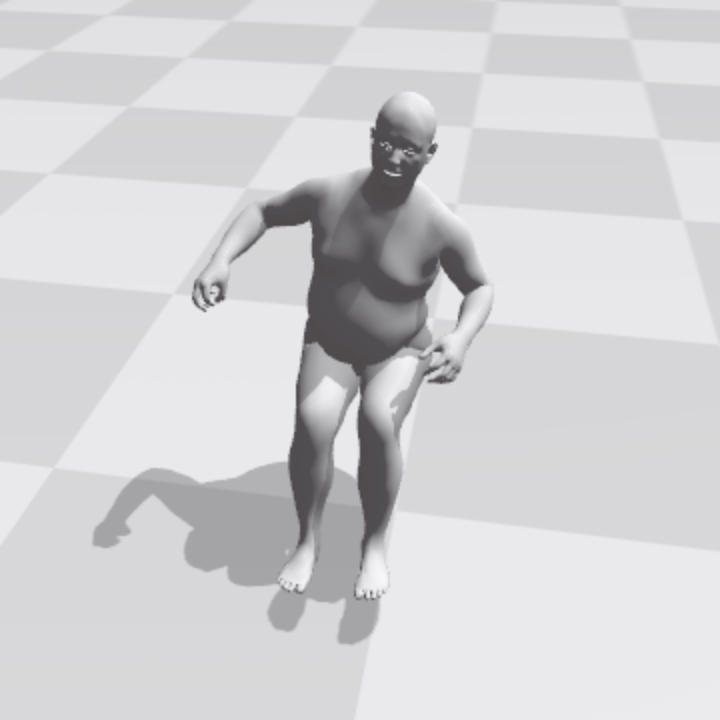}
    \includegraphics[width=0.19\textwidth]{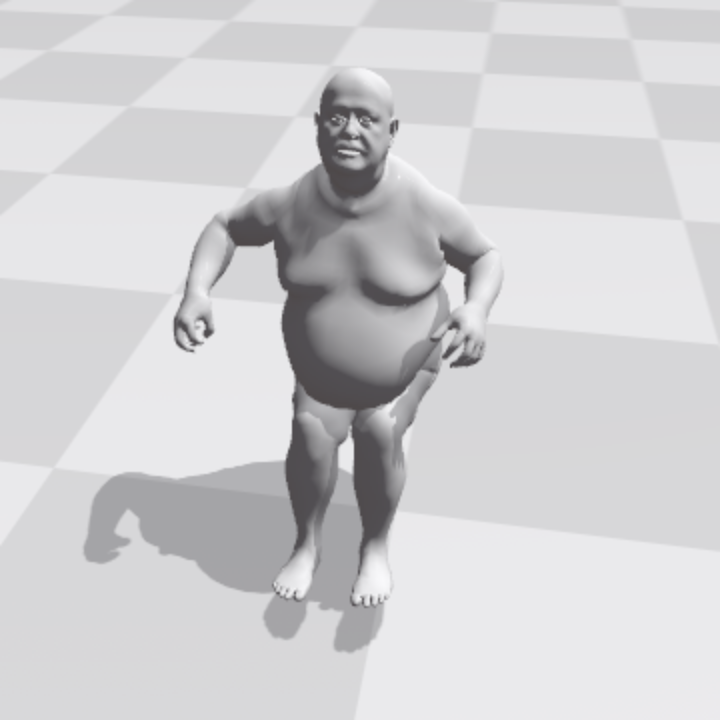}
    \includegraphics[width=0.19\textwidth]{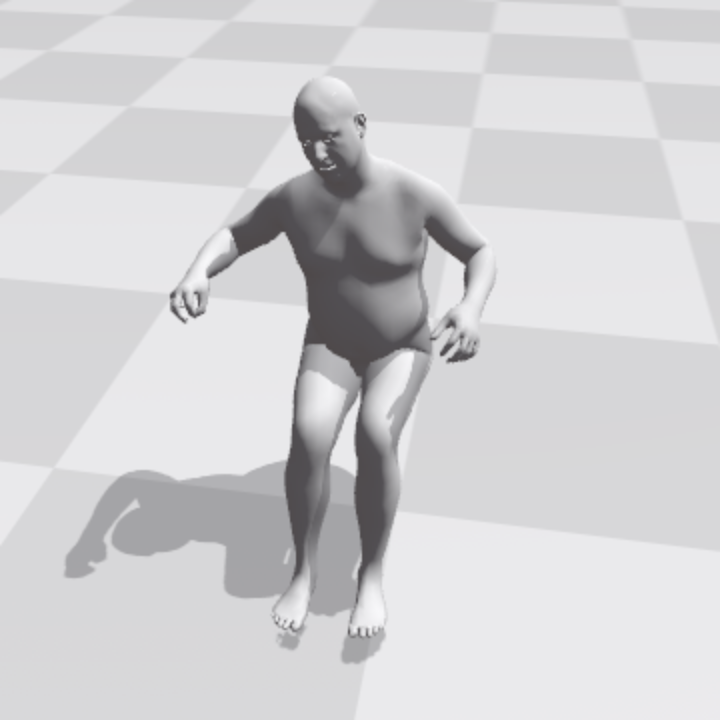}
    \includegraphics[width=0.19\textwidth]{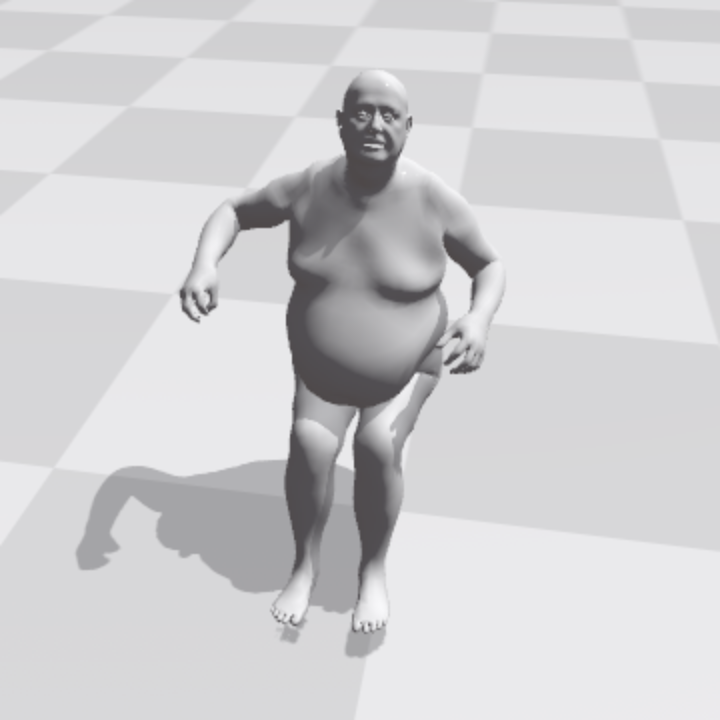}
    \includegraphics[width=0.19\textwidth]{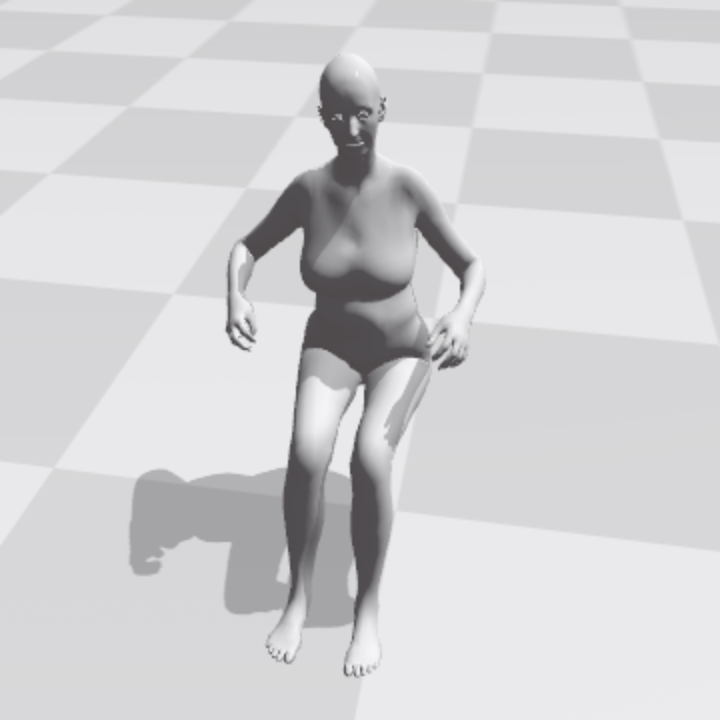}
    \includegraphics[width=0.19\textwidth]{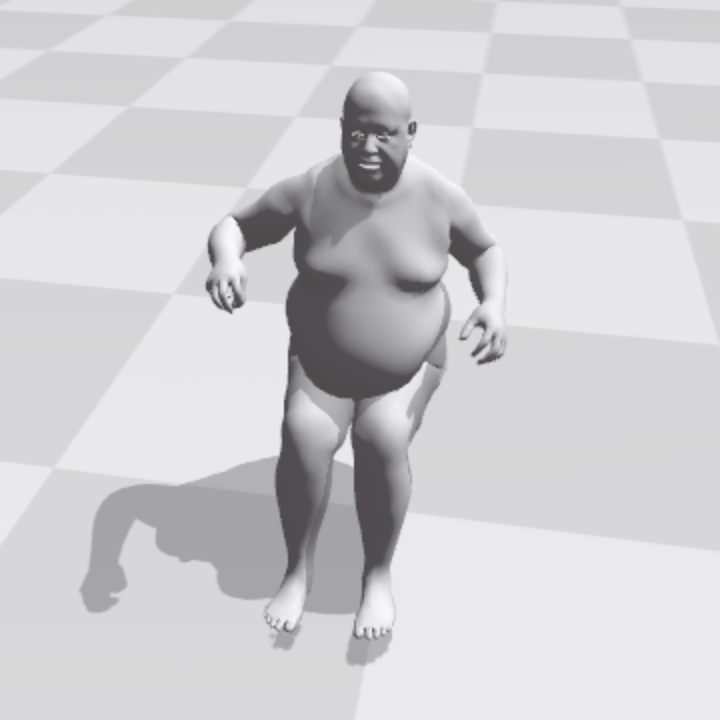}
    \includegraphics[width=0.19\textwidth]{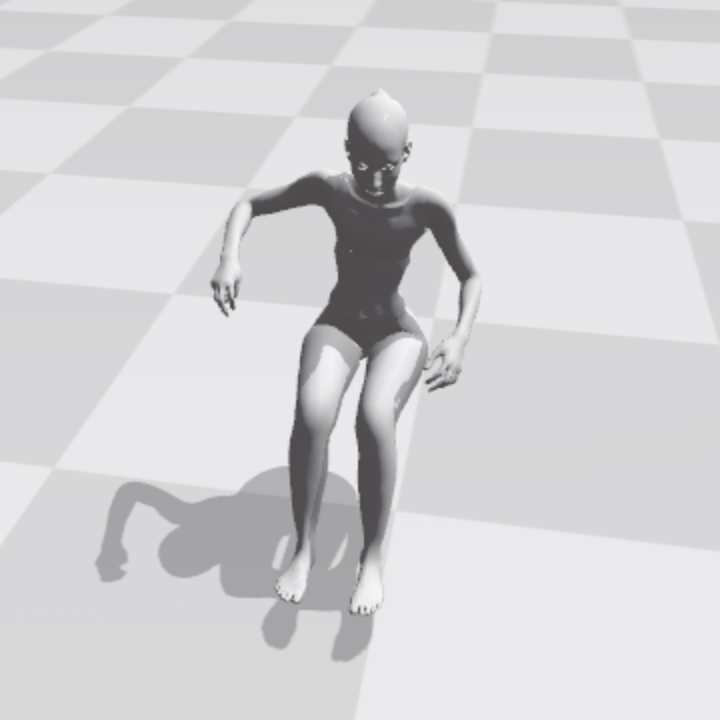}
    \includegraphics[width=0.19\textwidth]{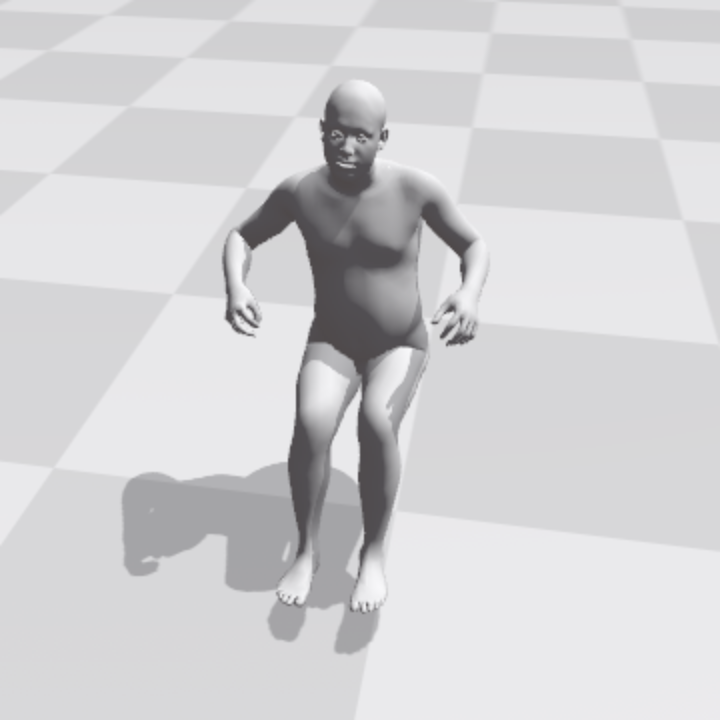}

    \caption{Some example human meshes we generate for simulation training. From top left to bottom right: a mesh from arm pose sub-range 1 to a mesh from arm pose sub-range 27.}
    \label{fig:human_mesh}
\end{figure}

\section{Simulation Experiments}
\subsection{Experimental Setup}
\subsubsection{Human Mesh Generation}
We use SMPL-X~\cite{SMPL-X:2019} model to generate human meshes of different body sizes and shapes. The body shape of the mesh is controlled by a 10D latent vector $\beta$. We uniformly sampled $\beta$ from $[-2, 5]$. The height of the mesh is uniformly sampled from $[1.5, 1.9]$. The gender is uniformly sampled from [Male, Female]. Some example generated meshes are shown in \figurename~\ref{fig:human_mesh}. 

\subsubsection{Simulation Garments}
We use a hospital gown and 4 cardigans chosen from the Cloth3D~\cite{bertiche2020cloth3d} dataset. 
These garments have different geometries and are shown in \figurename~\ref{fig:garments}. We scale the garment mesh so they are of proper size for dressing. 

\subsubsection{Arm Pose Range Decomposition}
For the shoulder joint $\phi_1$, we decompose the full interval $[-20, 30]$ to 3 intervals $[-20, -8], [-8, 18], [18, 30]$. For the inwards-outwards elbow joint angle $\phi_2$, we decompose the full interval $[-20, 20]$ to 3 intervals $[-20, -8], [-8, 8], [8, 20]$. For the upwards-downwards joint angle $\phi_3$, we decompose the full interval $[-20, 30]$ to 3 intervals $[-20, -3], [-3, 14], [14, 30]$. The cross product of all of these intervals results in 27 pose sub-ranges. \figurename~\ref{fig:human_mesh} shows an example mesh from each of the pose sub-ranges. 

\subsubsection{Simulator Parameters}
We use SoftGym~\cite{lin2021softgym} based on the NVIDIA FleX simulator. FleX simulates cloth as a collection of particles connected by springs. We set the stiffness of the stretch, bend, and shear spring connections to 0.3, 0.3, 0.3, respectively. The particle radius is 0.625 cm. The particle, dynamic, and static fricition coefficient are set to be 0.3, 0.3, and 0.3, respectively.

\subsection{Baseline Implementation}
\subsubsection{Direct Vector} 
For the policy, we use a classification-type PointNet++.  We use 2 set abstraction layers, followed by a global max pooling layer, and a MLP. The set abstraction radius are [0.05, 0.1], and the sampling ratios are [1, 1].  The final MLP is of size $[256, 256, 256, 128, 128, 128, 128, 128, 128]$. Since the Direct Vector policy does not have the feature propogation layers, the final MLP is set to be larger so the total number of parameters for the Dense Transformation and the Direct Vector policy is similar.  We train Direct Vector policy using the same SAC parameters as training the Dense Transformation Policy. 

\subsubsection{TD-MPC} 
We use the official implementation from the authors\footnote{https://github.com/nicklashansen/tdmpc}. For the MPPI planning, we use 8 iterations with 512 samples per iteration. The number of elites is 64, and planning horizon is 5. 
 We use a classification-type PointNet++ (the same as the one used for Direct Vector policy) to encode the partial point cloud into a latent vector.
The dimension of the latent vector is 100. Other hyper-parameters are set to the default values as in TD-MPC.

\subsubsection{PCGrad} We use a re-implementation of PCGrad~\cite{yu2020gradient} in PyTorch~\footnote{https://github.com/WeiChengTseng/Pytorch-PCGrad}. The dressing task can be viewed as multi-task learning where each arm pose sub-range is treated as a separate task. As we have $27$ arm pose sub-ranges, we keep $27$ replay buffers, one for each sub-range. We also follow~\cite{yu2020gradient} to learn an individual entropy temperature $\alpha$ in SAC for each task. We wrap the Adam~\cite{kingma2014adam} optimizer with PCGrad. In each training step, we randomly sample $16$ replay buffers. For each of the sampled replay buffer, we randomly sample $4$ transition tuples, $\{o_n, a_n, r_n, o'_n\}_{n=1}^4$,  to form a stratified batch of size $64$, and use PCGrad to perform the gradient update.

\subsubsection{Deep Haptic MPC}
We re-implemented Deep Haptic MPC \cite{erickson2018deep} for our environment. We train an one-step force prediction model $F(\bm{x}_{1:t}, \bm{a}_{t+1})$ given the robot end-effector state history $\bm{x}_{1:t}$ and action $\bm{a}_{t+1}$ following the original paper. The end-effector state measurements $x_t=(\bm{\rho}, \bm{v}, f) \in \mathbb{R}^{13}$ at time $t$ include the 6D position $\bm{\rho}$, 6D velocity $\bm{v}$, and force $f$ applied at the robot's end-effector. We do not use any visual information to train the force prediction model. During test time, we define a cost function that encourages lower force applied to the human arm and higher dressing performance. Since the original paper assumes a single fixed arm pose, sampling actions whose velocity lies within a hemisphere facing the $+x$ global
coordinate axis will encourage task progression. Different from \cite{erickson2018deep}, we have a diverse set of arm poses. Therefore, we use the position of the human hand, elbow, and shoulder to guide the action sampling. At each time step, if the garment is on the forearm, we sample $N=128$ candidate actions whose dot product with the direction from the human's finger to the elbow is positive; if the garment is on the upper arm, we sample $N=128$ candidate actions whose dot product with the direction from the human's elbow to the shoulder is positive. We always sample candidate actions that point to the direction of task progression. The input to the cost function includes a history of robot end-effector's measurements $\bm{x}_{1:t}$ and the next candidate action sample $\bm{a}_{t+1}$. The cost function penalizes actions that lead to large forces and actions whose magnitude is large. It also encourages actions that move forward along the human arm. Formally, the cost function is represented by three weighted terms as follows: 
\begin{align*}
    J(\bm{x}_{1:t}, \bm{a}_{t+1}) ={}& w_1\  |F(\bm{x}_{1:t}, \bm{a}_{t+1}|_1| \\ 
                                     & - w_2\ \bar{\bm{d}} \cdot \bm{a}_{t+1}^{\text{translation}} \\
                                     & + w_3\  ||\bm{a}_{t+1}||^2_2
\end{align*},
where $\bm{d}$ is the vector from human's finger to elbow when the garment is still on forearm and from human's elbow to shoulder when the garment is on the upper arm. We run this baseline on three selected pose regions. For each region, 
we collect training trajectories on the first 45 poses and 5 garments to train the force prediction model and we report the evaluation results on the last 5 poses in Table I in the paper. This baseline lacks the ability to generalize to diverse poses since it does not take in as input the visual observation of the arm and the garment. We find it perform worse than our proposed method on all three selected pose regions. 

\subsubsection{Heuristic Motion Planning}
As mentioned in the main paper, the motion planning baseline finds a collision-free robot end-effector path along the human arm. The following constraints are used for planning the path. First, the normal of the grasping tool (mounted on the robot end-effector) should align with the normal of the arm plane, which is the plane defined by the person’s finger, elbow, and shoulder points. Second, the forward direction of the end-effector should align with the direction of the forearm (the line connecting finger and elbow points) or the direction of the upper arm (the line connecting the elbow and shoulder points). Third, the path should go through three way points, which are defined as points above a certain distance from the finger, elbow, and shoulder points. Forth, the path needs to have no collision with the human arm. 
This baseline's average upper arm dressed ratio is 0.32 on all 27 arm pose regions, compared to 0.68 with our method (Table II in the paper).
We find such a method to perform well when there is no sharp bending around the shoulder and the elbow, and worse when the bends are sharp, aligning with findings in prior work~\cite{kapusta2019personalized}.

\section{Real-world Experiments}
\subsection{Human Study Procedure}
\noindent\textbf{Human arm capture. } To capture only the human arm point cloud, we manually select 3 pixels on the depth image that correspond to the shoulder, elbow, and hand point of the participant (this step can be replaced by using a human pose estimator in future work). We then transform these three pixels to three points in the robot frame. We form two lines: the first connects the hand point and the elbow point (forming a line along the forearm), and the second connect the elbow point and the shoulder point (forming a line along the upper arm). Then, we crop the point cloud by only keeping points whose distances to either of these two lines are smaller than a threshold $\Delta$. We set $\Delta = 7.5$ cm in our experiment. 

\noindent\textbf{Dressing time cost.}
Each dressing trial lasts between 1 to 2 minutes. It takes roughly 1 minute (as mentioned above) to perform the arm cropping and moving the Sawyer to be near the participant's hand. The actual dressing time is roughly between 40 seconds and 80 seconds. It usually takes 1 hour to 1.5 hours to finish the 30 dressing trials for each participant (including all other time cost such as rest time for the participant).

\noindent\textbf{Scripts for participant.}
Before the study begins, we read and show the following script to the participant for him/her to get familiar with the study procedure:

\textit{We are conducting a study to evaluate a robot-dressing system. The robot will dress the garment on your right arm. 
We will now walk you through the steps we are taking. 
You will first read and sign the consent form, and fill a demographic form. 
We will then measure some statistics of your arm, including the forearm length, upper arm length, and the arm circumference before the study starts.
We will put a marker on your shoulder for the experiment. 
We will then start the study. There will be 30 dressing trials, 6 trials for 5 garments. Each trial will be on a different arm pose, and a different garment.
You will be asked to hold your arm static during the trial. We will show the arm pose you need to hold on a computer screen, and we will also demonstrate the pose by ourselves to you if it’s not clear.
After we tell you to start, you need to hold the pose static for 1 - 2 minutes while we operate the robot to perform the dressing. 
Very occasionally the robot gripper might have contact with you during the dressing process. If you feel uncomfortable during the trial, you can tell us to stop the trial any time. 
Occasionally there will be operation failures on us for a trial. We will repeat those trials when such failures happen. 
After each trial, please keep holding the arm while we measure some statistics of the dressing performance. Then you can rest the arm and you will be asked to fill a questionnaire. The questionnaire is a Likert item stating that “The robot successfully dressed the garment onto my arm”. 1 is strongly disagree and 7 is strongly agree. 
You can rest whenever you feel tired, just let us know. 
After 6 trials for each garment, we will change the garment and you can also rest.    
}

\begin{figure}
    \centering
    \includegraphics[width=0.32\textwidth]{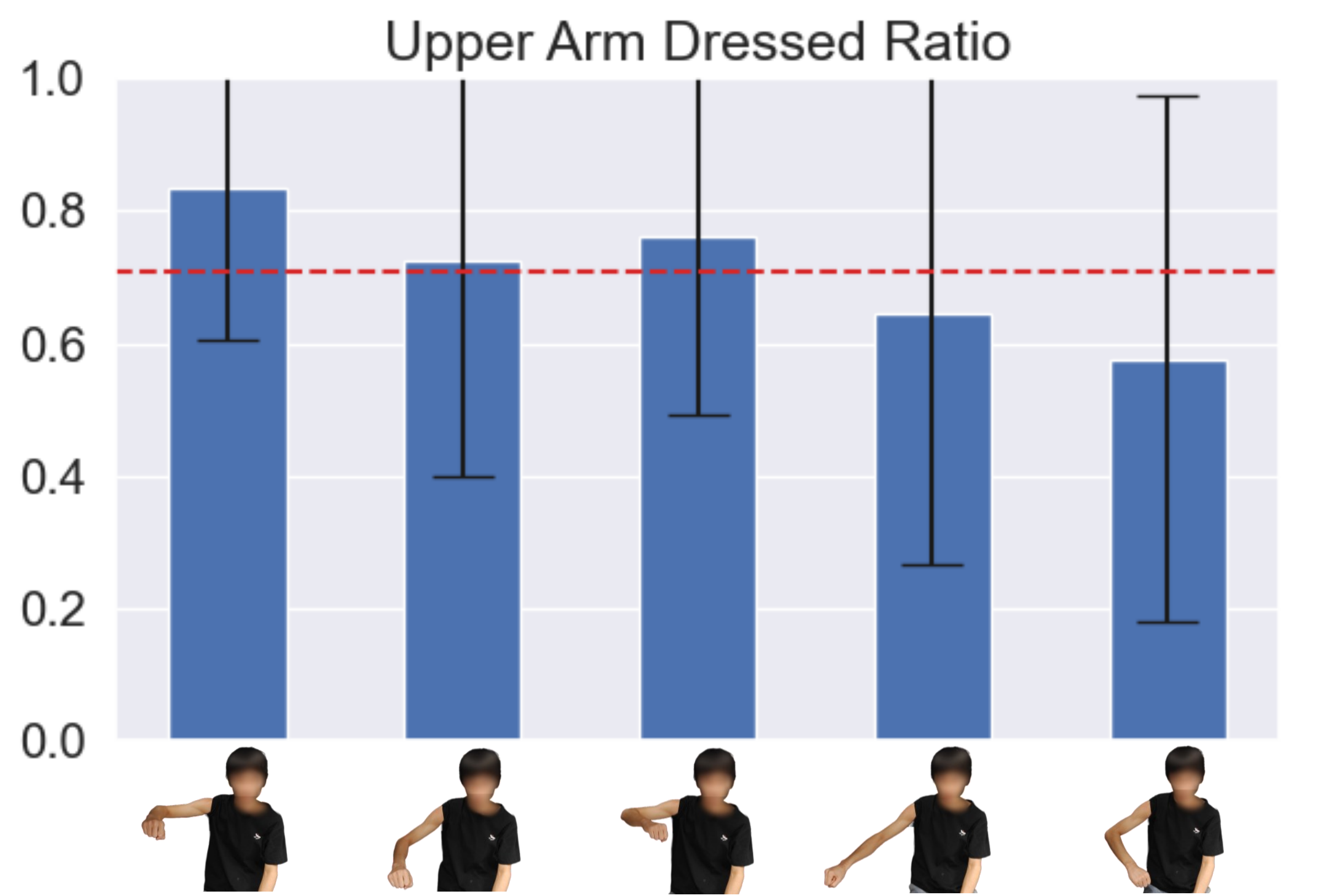}
    \includegraphics[width=0.32\textwidth]{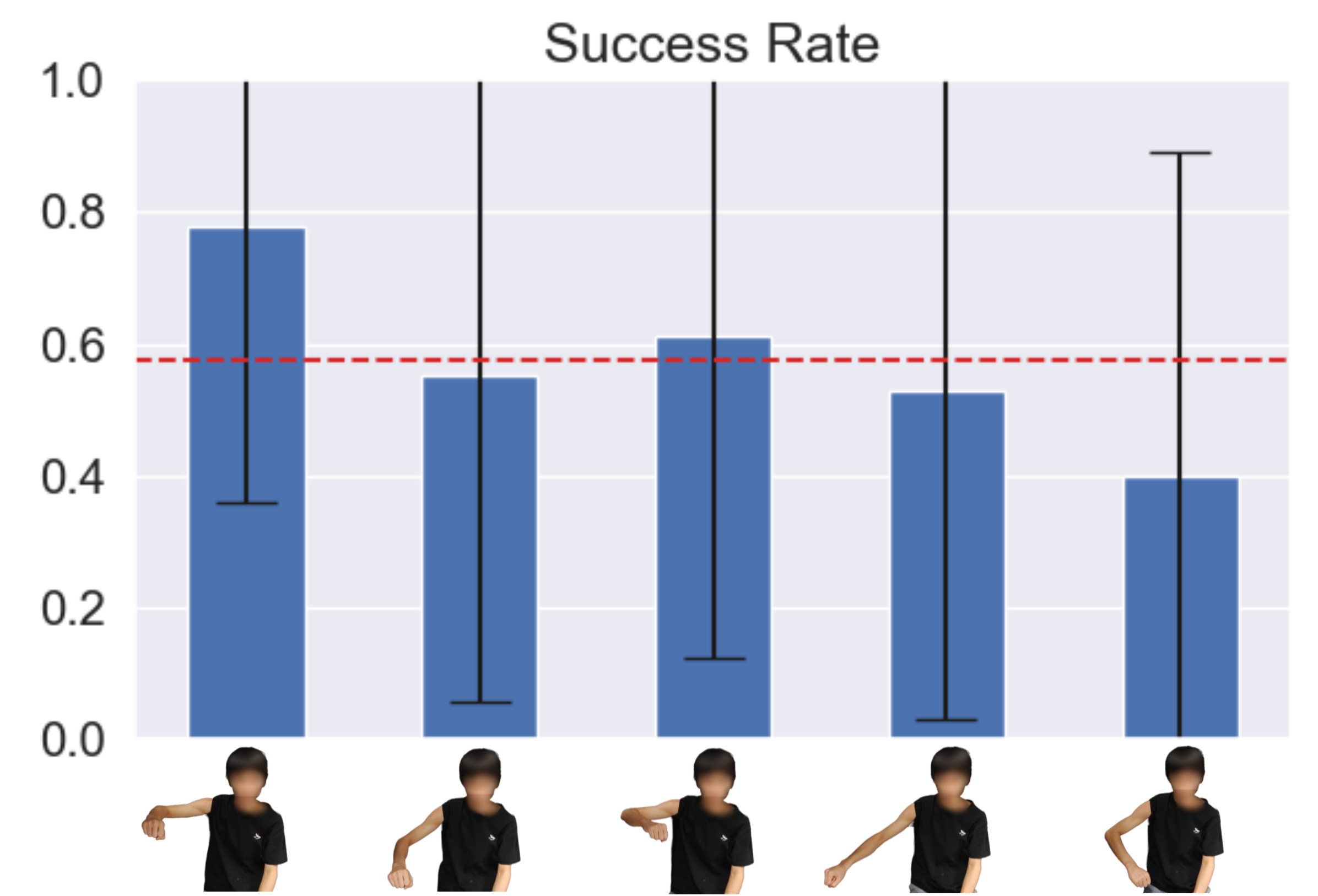}
    \includegraphics[width=0.32\textwidth]{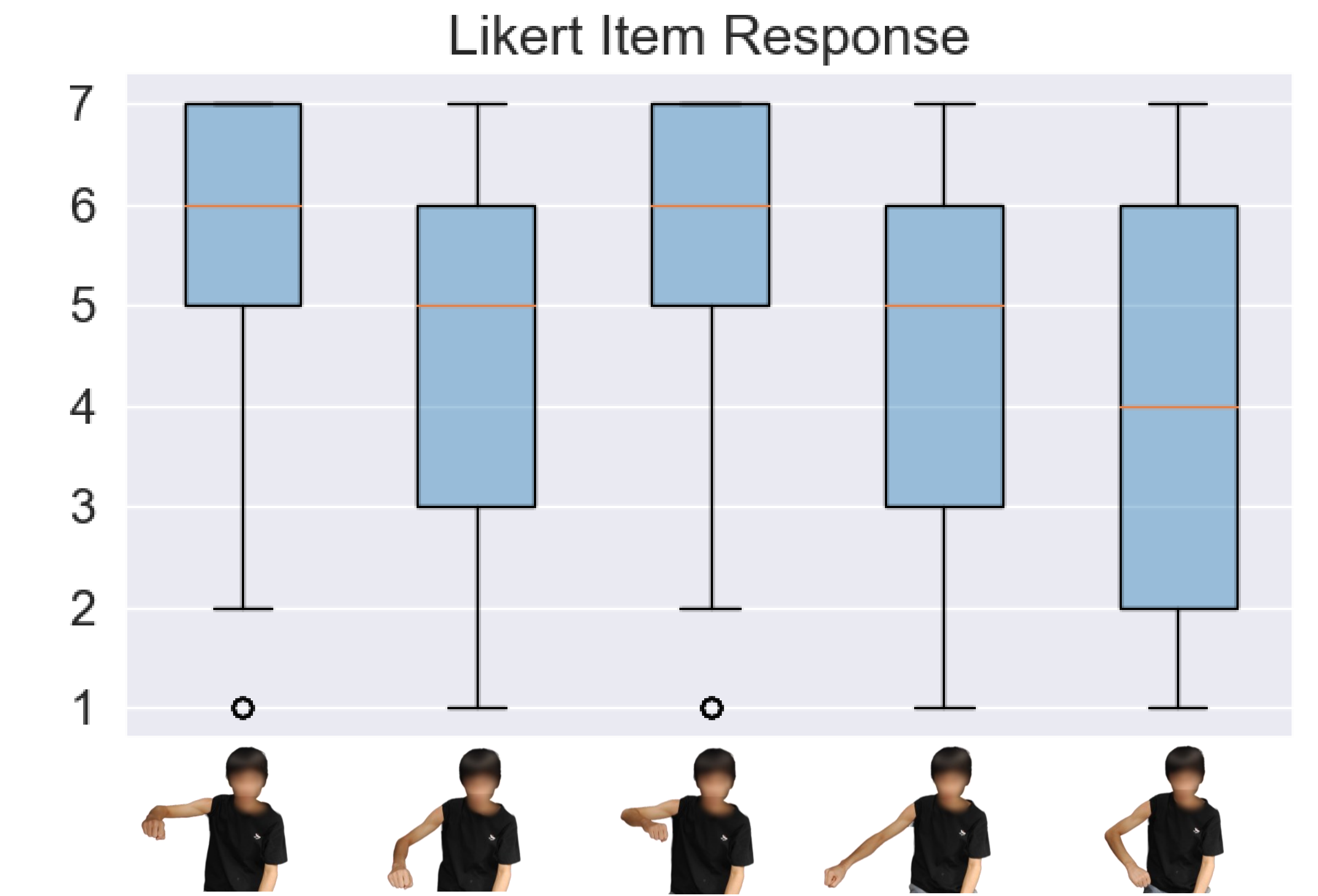}
    \caption{Upper arm dressed ratio, success rate, and Likert item responses for different poses in the human study. The results are averaged over all 425 dressing trials from 17 participants in the human study.}
    \label{fig:pose_performance_supplement}
\end{figure}

\subsection{More Real-world Experimental Results}
We present the upper arm dressed ratio, success rate, and Likert item responses for each of the poses we tested in the human study in \figurename~\ref{fig:pose_performance_supplement}. Again, we notice that some poses (the 5th one) is harder than others. 

\begin{figure}
    \centering
    \includegraphics[width=0.25\textwidth]{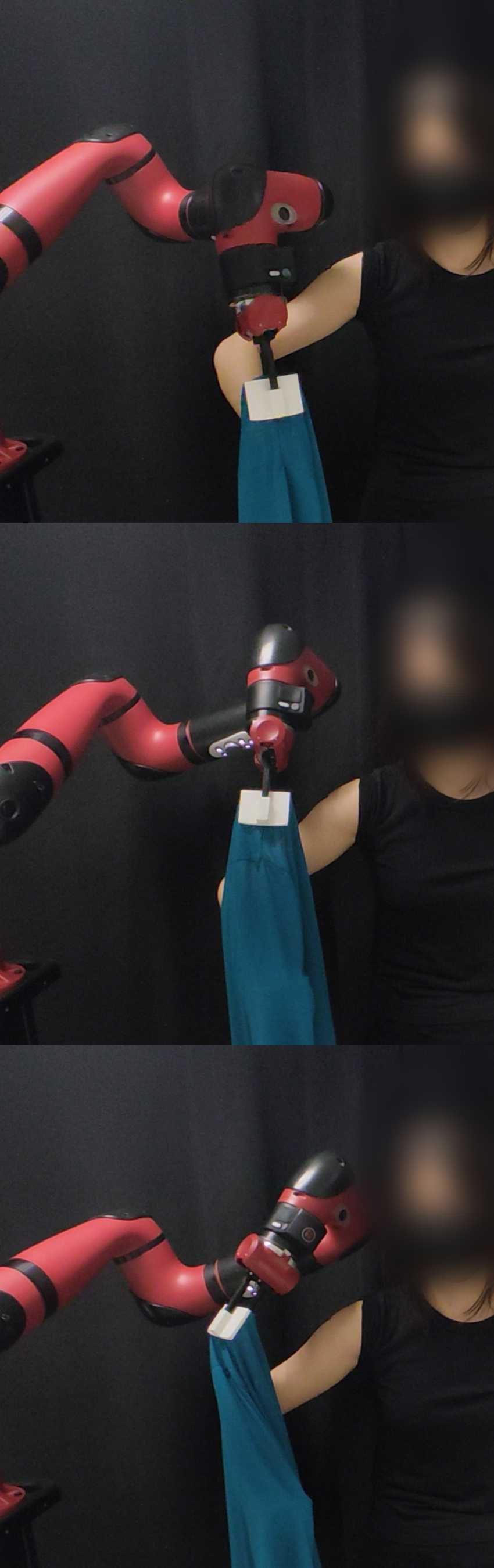}
    ~~
    \includegraphics[width=0.25\textwidth]{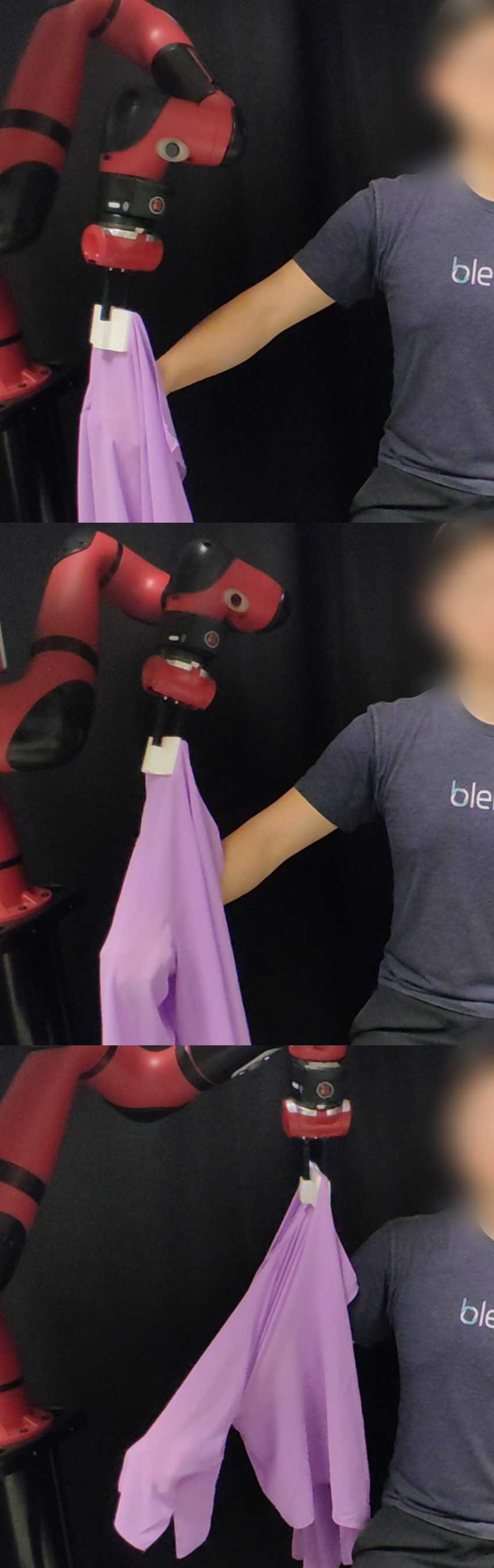}
    \caption{Failure cases of our system in the human study. Left: the policy gets stuck and stops output actions that make any further progress of  the task. Right: the garment gets caught on the participant because the policy actions pull the gripper to high above the participant's arm. }
    \label{fig:failure}
\end{figure}

\begin{figure}
    \centering
    \includegraphics[width=0.6\textwidth]{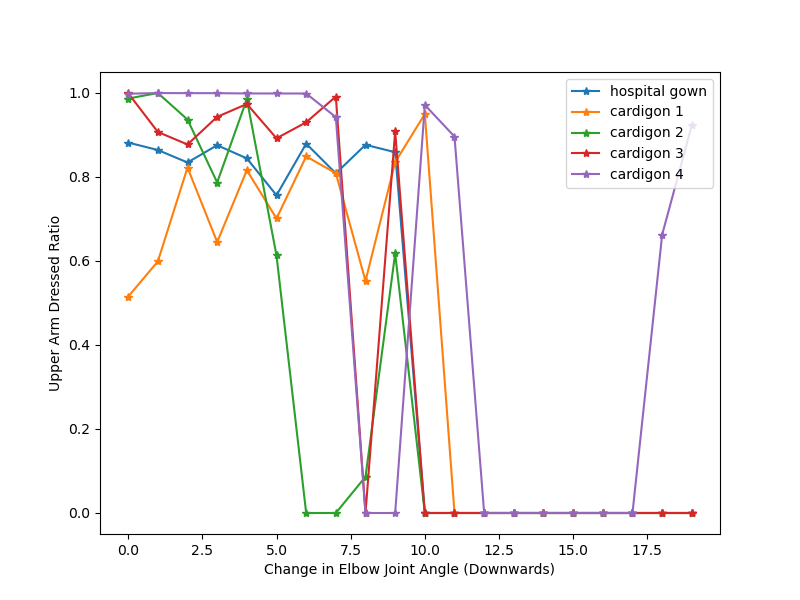}
    \includegraphics[width=0.6\textwidth]{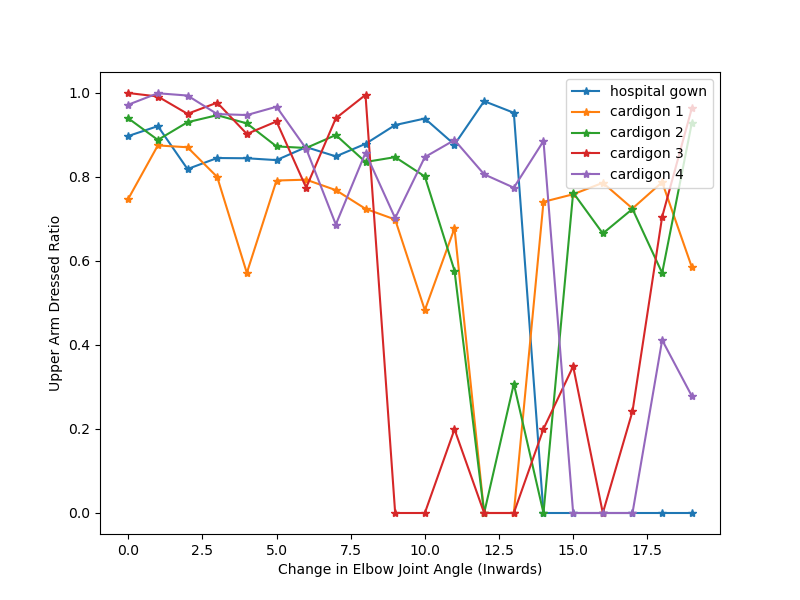}
    \includegraphics[width=0.6\textwidth]{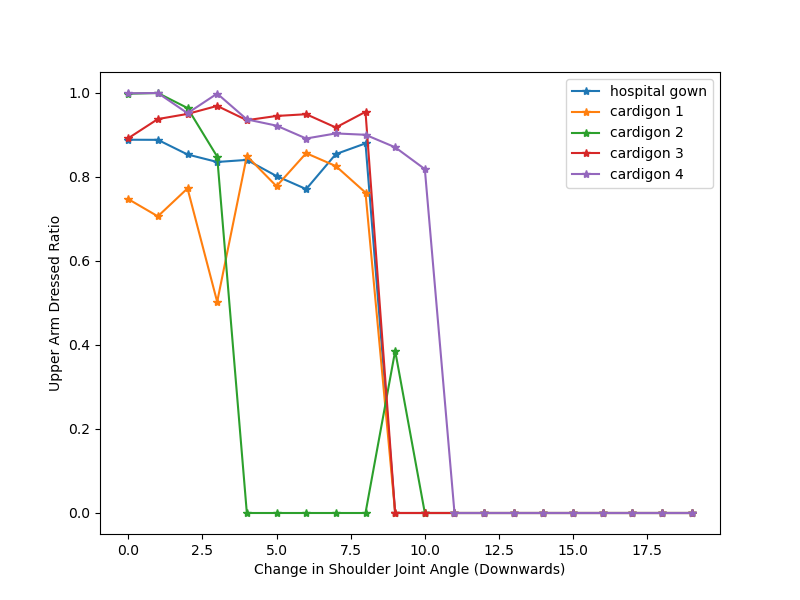}
    
    \caption{Upper arm dressed ratio versus changes in arm joint angles.}
    \label{fig:joint_angle_change}
\end{figure}

\subsection{Failure Cases}
\figurename~\ref{fig:failure} shows two failure cases from our policy. The first failure case is that the policy gets stuck and cannot output actions that make further progress for the task, e.g., the policy oscillates between moving up and down and does not move forward. This might be due to the gap in sim2real transfer or out-of-distribution arm poses/sizes/shapes in the real world. The second failure case occurs when the garment gets caught on the participant. Since it is visually difficult to detect if the garment has gotten stuck, incorporating force-torque sensing could be helpful in addressing this failure case. 

\section{Out of Distribution Evaluation}
As mentioned in the main paper, we perform evaluation of our system when the static arm assumption is relaxed. In simulation, we evaluate how our system performs if the participants change their shoulder or elbow joint angles after we capture the initial arm point cloud, on 1 pose sub-region and 5 garments. We test 3 types of joint angle changes: lowering down the shoulder joint, lowering down the elbow joint, and bending inwards the elbow joint. Figure~\ref{fig:joint_angle_change} demonstrates how the upper arm dressed ratio varies as the change of joint angle varies. Overall, we find our system to be robust to 8.6 degrees of change in shoulder and elbow joint angles (averaged across 3 types of joint angle changes and 5 garments) while maintaining 75\% of the original performance.

\end{appendices}



\end{document}